\RequirePackage{amsmath} 
\documentclass[runningheads]{llncs}

\usepackage[T1]{fontenc}
\usepackage{csquotes}
\usepackage{amssymb}

\usepackage{graphicx}
\usepackage{svg}
\usepackage{xcolor}

\usepackage{tikz}
\usepackage{tikz-3dplot}
\usepackage{pgfplots}
\pgfplotsset{compat=1.16}
\usetikzlibrary{calc}

\usepackage{subcaption}
\usepackage{booktabs}
\usepackage{array}
\usepackage{enumitem}

\usepackage[acronym]{glossaries}
\usepackage{hyperref}  
\hypersetup{
    colorlinks=true,
    linkcolor=black,
    urlcolor=blue,
    citecolor=blue,
    hidelinks
}
\usepackage[capitalize]{cleveref}  
\usepackage[sorting=none]{biblatex}
\makeglossaries

\newglossaryentry{IE}{
    name={IE},
    first={Information Engine (IE)},
    description={Information Engine}
}

\newglossaryentry{OPCUA}{
    name={OPC UA},
    first={Open Platform Communications Unified Architecture (OPC UA)},
    description={Open Platform Communications Unified Architecture}
}

\newglossaryentry{MBSE}{
    name={MBSE},
    first={Model-Based Systems Engineering (MBSE)},
    description={Model-Based Systems Engineering}
}

\newglossaryentry{SysML}{
    name={SysML},
    first={Systems Modeling Language (SysML)},
    description={Systems Modeling Language}
}

\newglossaryentry{SLURM}{
    name={SLURM},
    first={Simple Linux Utility for Resource Management (SLURM)},
    description={Simple Linux Utility for Resource Management}
}

\newglossaryentry{AAS}{
    name={AAS},
    first={Asset Administration Shell (AAS)},
    description={Asset Administration Shell}
}

\newglossaryentry{DSL}{
    name={DSL},
    first={Domain-Specific Language (DSL)},
    description={Domain-Specific Language}
}
\newglossaryentry{ml}{
    name={ML},
    first={Machine Learning (ML)},
    description={Machine Learning}
}

\newglossaryentry{AI}{
    name={AI},
    first={Artificial Intelligence (AI)},
    description={Artificial Intelligence}
}

\newglossaryentry{MLOps}{
    name={MLOps},
    first={Machine Learning Operations (MLOps)},
    description={Machine Learning Operations}
}

\newglossaryentry{devops}{
    name={DevOps},
    first={Development Operations (DevOps)},
    description={Development Operations}
}

\newglossaryentry{OT}{
    name={OT},
    first={Operational Technology (OT)},
    description={Operational Technology}
}

\newglossaryentry{IT}{
    name={IT},
    first={Information Technology (IT)},
    description={Information Technology}
}

\newglossaryentry{CI}{
    name={CI},
    first={Continuous Integration (CI)},
    description={Continuous Integration}
}

\newglossaryentry{CD}{
    name={CD},
    first={Continuous Delivery (CD)},
    description={Continuous Delivery}
}

\newglossaryentry{R4}{
    name={RAMI 4.0},
    first={Reference Architectural Model for Industry 4.0 (RAMI 4.0)},
    description={Reference Architectural Model for Industry 4.0}
}

\newglossaryentry{I4}{
    name={Industry 4.0},
    first={Industry 4.0},
    description={Industry 4.0}
}

\newglossaryentry{I95}{
    name={ISA-95},
    first={International Society of Automation Standard 95 (ISA-95)},
    description={ISA-95}
}

\newglossaryentry{MES}{
    name={MES},
    first={Manufacturing Execution Systems (MES)},
    description={Manufacturing Execution Systems}
}

\newglossaryentry{ERP}{
    name={ERP},
    first={Enterprise Resource Planning (ERP)},
    description={Enterprise Resource Planning}
}

\newglossaryentry{SCADA}{
    name={SCADA},
    first={Supervisory Control and Data Acquisition (SCADA)},
    description={Supervisory Control and Data Acquisition}
}

\newglossaryentry{ICS}{
    name={ICS},
    first={Industrial Control Systems},
    description={Industrial Control Systems}
}

\newif\ifAppendix
\Appendixfalse

\newif\ifBlindReview
\BlindReviewfalse

\newif\ifNotBlindReview
\ifBlindReview\NotBlindReviewfalse\else\NotBlindReviewtrue\fi

\newif\ifShowComments
\ShowCommentsfalse

\newcommand{\sthu}[1]{\ifShowComments\marginpar{\tiny\textcolor{red}{StHu: #1}}\fi}

\begin{document}

\title{Embedding the MLOps Lifecycle into OT Reference Models}
\ifBlindReview
	\author{Anonymous Author(s)}
	\authorrunning{Anonymous}
	\institute{Anonymous Institution\\
		Anonymous Location\\
		\email{anonymous@example.com}}
\else
	\author{Simon Schindler \inst{1} \and Christoph Binder\inst{2} \\ Lukas Lürzer\inst{1} \and Stefan Huber\inst{1}}
	\authorrunning{S.~Schindler et al.}
	\institute{$^1$Josef Ressel Centre for Intelligent and Secure Industrial Automation\\
		$^2$Josef Ressel Centre for Dependable System-of-Systems Engineering\\
		Salzburg University of Applied Sciences, Austria \\
		\email{\{simon.schindler,christoph.binder,lukas.luerzer,stefan.huber\}\\@fh-salzburg.ac.at}}
\fi
\maketitle              

\begin{abstract}
	\gls{MLOps} practices are increasingly
	adopted in industrial settings, yet their integration with
	\gls{OT} systems presents significant challenges. This paper
	analyzes the fundamental obstacles in combining \gls{MLOps} with \gls{OT} environments
	and proposes a systematic approach to embed \gls{MLOps} practices into
	established \gls{OT} reference models. We evaluate the suitability of the \gls{R4}
	and the \gls{I95} for \gls{MLOps} integration and present a detailed mapping of \gls{MLOps}
	lifecycle components to \gls{R4} exemplified by a real-world use case.
	Our findings demonstrate that
	while standard \gls{MLOps} practices cannot be directly transplanted to \gls{OT}
	environments, structured adaptation using existing reference models can
	provide a pathway for successful integration.
\end{abstract}

\section{Introduction}
\subsubsection{Motivation and Problem Statement}
The increasing adoption of \gls{ml} in industrial environments promises
significant improvements in efficiency, quality, and productivity. However, the
systematic integration of \gls{MLOps} practices into \gls{OT} systems remains a
significant challenge. While \gls{MLOps} has proven successful in traditional
\gls{IT} environments, its application in industrial settings is hindered by
strict requirements for safety, real-time performance, and reliability.

Industrial automation systems face a fundamental conflict when adopting
\gls{MLOps} practices: they must maintain strict operational requirements while
accommodating the dynamic nature of \gls{ml} systems. Traditional
\gls{OT} systems are designed for stability and deterministic behavior, whereas
\gls{MLOps} emphasizes continuous updates and iterative improvement. This
creates tensions in several key areas. Real-time performance requirements
conflict with the variable latency typical of \gls{ml} inference. Safety and
reliability standards restrict the ability to perform continuous deployment.
Resource constraints on edge devices limit the complexity of deployable models.
Additionally, legacy equipment and protocols complicate data collection and
model deployment.
%

\subsubsection{Research Questions.}
These challenges necessitate a systematic approach to adapt \gls{MLOps}
practices for industrial environments while preserving the core principles of
\gls{OT} systems. The goal is to bring agile and iterative \gls{ml} practices
to manufacturing while maintaining operational stability. This paper examines
how \gls{MLOps} can be effectively integrated into established \gls{OT}
reference models by answering the following research questions:

\begin{enumerate}
	\item What are the main challenges for the integration of
	      \gls{MLOps} into \gls{OT} systems?
	\item Given these challenges, can prevalent
	      \gls{MLOps} reference models be embedded into prevalent \gls{OT} reference
	      architecture models like \gls{R4} or \gls{I95}?
\end{enumerate}

\subsubsection{Related Work} To the best of our knowledge, there is no existing
work that systematically embeds the \gls{MLOps} lifecycle into \gls{OT}
reference models. However, there are works that address the challenges of
\gls{MLOps} in industrial settings and propose concrete solutions that might be
relevant for the reader of this paper. In~\cite{Faubel2023MLOps4.0}, Faubel et
al.\ provide a systematic analysis of \gls{MLOps} challenges in \gls{I4}
environments. Their findings reveal that while most \gls{MLOps} activities
remain similar to traditional software contexts, significant challenges emerge
in data collection, infrastructure management, and deployment. While they
identify these challenges, they do not provide a comprehensive framework that
integrates \gls{MLOps} within established \gls{OT} reference models.

In~\cite{Hegedus2023TailoringNeeds}, Hegedüs et al.\ propose concrete solutions
to these industrial \gls{MLOps} challenges through a specialized engineering
toolchain. Their architecture addresses key requirements of industrial
environments: digital twin integration for testing and validation,
infrastructure adaptation for edge deployment, robust monitoring capabilities,
and sandboxes for development and validation of \gls{ml} models. However, their
approach focuses primarily on technical implementation rather than establishing
a standardized methodology that aligns with existing \gls{OT} frameworks.

This research gap - the lack of a systematic integration of \gls{MLOps}
practices within established \gls{OT} reference models - presents significant
challenges for organizations attempting to implement \gls{ml} solutions in
industrial environments. Our work addresses this gap by providing a
comprehensive framework that embeds the \gls{MLOps} lifecycle into \gls{OT}
reference models, thereby creating a standardized approach that facilitates
adoption across industrial sectors.

\sthu{Can we conclude, what the gap left open is? Maybe merged in a summary
	conclusion with above, if the conclusion is the same.}

\sthu{It is important to make the research gap explicit so our contribution
	becomes weight.}

\subsubsection{Contribution.}
The main contributions of this work are:
\begin{itemize}
	\item The identification of key challenges for integrating \gls{MLOps} into
	      \gls{OT} environments, focusing on embedding respective \gls{MLOps}
	      phases into appropriate layers of the \gls{OT} architecture - extending
	      beyond the analysis of Faubel et al.~\cite{Faubel2023MLOps4.0} by
	      providing concrete integration patterns.
	\item The systematic evaluation of \gls{OT} reference models \gls{R4} and
	      \gls{I95} for their suitability to integrate \gls{MLOps}, addressing
	      the gap left by Hegedüs et al.~\cite{Hegedus2023TailoringNeeds} who
	      focus on technical implementation without reference model integration.
	\item A comprehensive mapping of the \gls{MLOps} lifecycle into the
	      \gls{R4} reference model.
	\item A validation of the proposed mapping through application to a
	      real-world industrial use case.
\end{itemize}

\section{Background}\label{sec:background}

\subsection{MLOps: Machine Learning Operations}\label{subsec:mlops}

The field of \gls{MLOps} is a collection of tools, techniques and
processes/models enabling structured approaches for the development, training,
evaluation, tracking, deployment and monitoring of \gls{ml} models. It adopts
and extends ideas taken from \gls{devops} for
the development and integration of \gls{ml} into software environments, while
addressing the specific challenges of operationalizing \gls{ml} systems
\cite{Kreuzberger2023MachineArchitecture,Salama2021practitioners}.

According to Symonidis et al.~\cite{Symonidis2022MLOpsDefinitions} and aligned with
the practitioners guide for \gls{MLOps}~\cite{Salama2021practitioners} by Google, the \gls{MLOps} lifecycle
comprises the following phases, as illustrated in Figure~\ref{fig:mlops}:
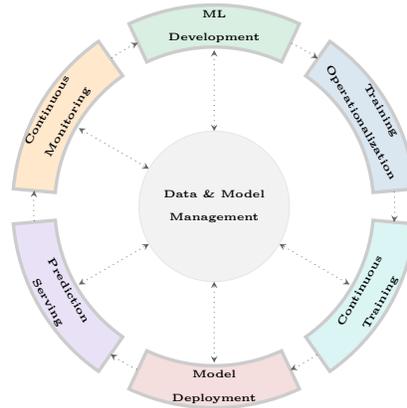
\begin{figure}[ht]
	\def\includingtikz{1}
	\begin{center}
		\scalebox{0.8}{
\ifdefined\includingtikz
\else
    \documentclass{standalone}
    \usepackage{tikz}
    \begin{document}
\fi

\begin{tikzpicture}
    \tikzset{
        inventory/.style={circle, minimum width=2.5cm, draw=black!10, fill=black!5, align=center, font=\footnotesize},
        phase/.style={text width=2.5cm, align=center, font=\footnotesize}
    }
    
    \definecolor{ml-blue}{RGB}{70,130,180}
    \definecolor{train-green}{RGB}{60,179,113}
    \definecolor{cont-orange}{RGB}{255,140,0}
    \definecolor{deploy-purple}{RGB}{147,112,219}
    \definecolor{pred-red}{RGB}{205,92,92}
    \definecolor{monitor-teal}{RGB}{72,209,204}
    
    \foreach \i in {-5,55,...,295} {
        \draw[<-,>=stealth,dotted, thin, black!60] (\i:3) arc (\i:\i+10:3);
    }
    
    \begin{scope}[shift={(0,0)}]
        \begin{scope}[rotate=5]
            \draw[ultra thick, fill=ml-blue, opacity=.2] (0:2.65) arc (0:50:2.65) -- (50:3.35) arc (50:0:3.35) -- cycle;
        \end{scope}
        \begin{scope}[rotate=65]
            \draw[ultra thick, fill=train-green, opacity=.2] (0:2.65) arc (0:50:2.65) -- (50:3.35) arc (50:0:3.35) -- cycle;
        \end{scope}
        \begin{scope}[rotate=125]
            \draw[ultra thick, fill=cont-orange, opacity=.2] (0:2.65) arc (0:50:2.65) -- (50:3.35) arc (50:0:3.35) -- cycle;
        \end{scope}
        \begin{scope}[rotate=185]
            \draw[ultra thick, fill=deploy-purple, opacity=.2] (0:2.65) arc (0:50:2.65) -- (50:3.35) arc (50:0:3.35) -- cycle;
        \end{scope}
        \begin{scope}[rotate=245]
            \draw[ultra thick, fill=pred-red, opacity=.2] (0:2.65) arc (0:50:2.65) -- (50:3.35) arc (50:0:3.35) -- cycle;
        \end{scope}
        \begin{scope}[rotate=305]
            \draw[ultra thick, fill=monitor-teal, opacity=.2] (0:2.65) arc (0:50:2.65) -- (50:3.35) arc (50:0:3.35) -- cycle;
        \end{scope}
    \end{scope}
    
    \foreach \angle/\startrad/\len in {90/2.6/1.25, 150/2.6/1.25, 210/2.6/1.25, 270/2.6/1.25, -30/2.6/1.25} {
        \draw[<->,>=stealth, dotted, thin, black!60] (\angle:\startrad) -- (\angle:\len);
    }
    
    \node[inventory] (inventory) at (0,0) {\tiny \textbf{Data \& Model}\\\tiny \textbf{Management}};
    
    \node[phase, rotate=0] at (0,3) {\tiny \textbf{ML \\ Development}};
    \node[phase, rotate=-60] at (2.6,1.5) {\tiny \textbf{Training}\\\tiny \textbf{Operationalization}};
    \node[phase, rotate=60] at (2.6,-1.5) {\tiny \textbf{Continuous\\ Training}};
    \node[phase, rotate=0] at (0,-3) {\tiny \textbf{Model \\ Deployment}};
    \node[phase, rotate=-60] at (-2.6,-1.5) {\tiny \textbf{Prediction\\Serving}};
    \node[phase, rotate=60] at (-2.6,1.5) {\tiny \textbf{Continuous Monitoring}};
\end{tikzpicture}

\ifdefined\includingtikz
\else
    \end{document}
\fi
		}
	\end{center}
	\caption{The \gls{MLOps} lifecycle as described in \cite{Salama2021practitioners} showing its key components and their interactions.}
	\label{fig:mlops}
	\undef\includingtikz
\end{figure}
\begin{itemize}
	\item \textbf{ML Development:} Development of reproducible procedures,
	      including data preparation, transformation, model training and evaluation.

	\item \textbf{Training Operationalization:} Automation of training pipeline deployment,
	      including packaging, testing, and validation processes.

	\item \textbf{Continuous Training:} Systematic execution of training pipelines triggered
	      by new data, code changes, or scheduled updates.

	\item \textbf{Model Deployment:} Deployment of models to serving environments for
	      experimental validation and production use.

	\item \textbf{Prediction Serving:} Production inference through deployed models.

	\item \textbf{Continuous Monitoring:} Assessment of model effectiveness and computational
	      efficiency in production environments.

	\item \textbf{Data and Model Management:} Central governance framework ensuring artifact
	      auditability, traceability, and compliance while facilitating asset reusability
	      and discovery.
\end{itemize}
As visualized in \cref{fig:mlops} the \textbf{Data and Model Management}
component is central to the lifecycle, bidirectionally interacting and
synchronizing with most phases. The interactions between the phases are mostly
unidirectional serving key outputs as inputs to the next phase. A high level
overview of the control flow is the following: Source code and data from the
\textbf{ML Development} phase are used to create the training pipeline in the
\textbf{Training Operationalization} phase. The training pipeline is then
executed in the \textbf{Continuous Training} phase, resulting in a trained
model which is then registered with \textbf{Data and Model Management} and
deployed in the \textbf{Model Deployment} phase. This model is then packaged in
a self contained unit and deployed to a serving environment in the
\textbf{Prediction Serving} phase which supplies logs about the model
performance to the \textbf{Continuous Monitoring} phase. Any feedback from the
monitoring phase is then used to update the model in the \textbf{ML
	Development} phase, which closes the cycle.

\subsection{Reference Models for Operational Technology}
Industrial automation systems rely heavily on \gls{OT},
which encompasses the hardware and software that monitors and controls physical
processes in industrial environments. Unlike \gls{IT}, which primarily manages
data-centric operations, \gls{OT} systems are cyber-physical systems and hence
directly interface with physical equipment and processes. Reference models,
such as \gls{R4} or the \gls{I95} standard, play a crucial role in structuring
and understanding these systems by providing standardized frameworks for system
organization, communication hierarchies, and security implementations. These
models have become particularly important as industries progress towards
greater integration of \gls{OT} and IT systems, enabling data-driven operations
while maintaining clear architectural boundaries and operational safety.
The following subsections provide a brief overview of the \gls{R4} and
\gls{I95} reference models.
\begin{figure}[t]
	\centering
	\begin{minipage}{0.4\textwidth}
		\centering
		\def\includingtikz{1}
		\scalebox{0.6}{  
			\hspace{-12.75cm}
			\ifdefined\includingtikz
    \ifdefined\highlightboxes
    \else
        \def\highlightboxes{}
    \fi
\else
    \documentclass{standalone}
    \usepackage{graphicx}
    \usepackage{xcolor}
    \usepackage{tikz}
    \usepackage{tikz-3dplot}
    \usepackage{pgfplots}
    \pgfplotsset{compat=1.16}
    \def\highlightboxes{
			}
\begin{document}
\fi

\ifdefined\ramiprefix
\else
	\def\ramiprefix{X}
\fi

\definecolor{maincolor}{RGB}{35, 95, 157}
\definecolor{secondcolor}{RGB}{137, 157, 35}
\definecolor{alertcolor}{RGB}{192, 83, 84}
\definecolor{darkshadecolor}{RGB}{147,112,219}
\colorlet{shadecolor}{darkshadecolor!10}
\colorlet{shadegrey}{black!5}
\definecolor{cLayer1}{RGB}{72,209,204}
\definecolor{cLayer2}{RGB}{60,179,113}
\definecolor{cLayer3}{RGB}{255,140,0}
\definecolor{cLayer4}{RGB}{147,112,219}
\definecolor{cLayer5}{RGB}{205,92,92}
\definecolor{cLayer6}{RGB}{70,130,180}

\usetikzlibrary{calc, shapes.symbols, shapes.misc, shapes.arrows, decorations,
	decorations.pathmorphing, fit, matrix, matrix.skeleton, backgrounds}
\tikzset{>=latex}

\tikzstyle{boxnodewhite}=[draw, rectangle, minimum height=2em]
\tikzstyle{boxnode}=[boxnodewhite, fill=shadecolor]
\tikzstyle{boxnodehalf}=[boxnode, dashed, opacity=0.5]

\makeatletter
\newdimen\multi@col@width
\newdimen\multi@col@margin
\newcount\multi@col@count
\multi@col@width=0pt

\tikzset{
	multicol/.code={%
			\global\multi@col@count=#1\relax
			\global\let\orig@pgfmatrixendcode=\pgfmatrixendcode
			\global\let\orig@pgfmatrixemptycode=\pgfmatrixemptycode
			\def\pgfmatrixendcode##1{\orig@pgfmatrixendcode%
				##1%
				\pgfutil@tempdima=\pgf@picmaxx
				\global\multi@col@margin=\pgf@picminx
				\advance\pgfutil@tempdima by -\pgf@picminx
				\divide\pgfutil@tempdima by #1\relax
				\global\multi@col@width=\pgfutil@tempdima
				\pgf@picmaxx=.5\multi@col@width
				\pgf@picminx=-.5\multi@col@width
				\global\pgf@picmaxx=\pgf@picmaxx
				\global\pgf@picminx=\pgf@picminx
				\gdef\multi@adjust@position{%
					\setbox\pgf@matrix@cell=\hbox\bgroup
					\hfil\hskip-1.5\multi@col@margin
					\hfil\hskip-.5\multi@col@width
					\box\pgf@matrix@cell
					\egroup
				}%
				\gdef\multi@temp{\aftergroup\multi@adjust@position}%
				\aftergroup\multi@temp
			}
			\gdef\pgfmatrixemptycode{%
				\orig@pgfmatrixemptycode
				\global\advance\multi@col@count by -1\relax
				\global\pgf@picmaxx=.5\multi@col@width
				\global\pgf@picminx=-.5\multi@col@width
				\ifnum\multi@col@count=1\relax
					\global\let\pgfmatrixemptycode=\orig@pgfmatrixemptycode
				\fi
			}
		}
}
\makeatother

\tdplotsetmaincoords{78}{50}
\begin{tikzpicture}[
		tdplot_main_coords,
		layer/.style={draw opacity=1.0, fill opacity=0.1},
		layer0/.style={layer, draw=cLayer1, fill=cLayer1},
		layer1/.style={layer, draw=cLayer2, fill=cLayer2},
		layer2/.style={layer, draw=cLayer3, fill=cLayer3},
		layer3/.style={layer, draw=cLayer4, fill=cLayer4},
		layer4/.style={layer, draw=cLayer5, fill=cLayer5},
		layer5/.style={layer, draw=cLayer6, fill=cLayer6}
	]
	\def\layerdy{0.8}
	\def\layerdz{0.5}
    \def\layerdx{1}

	\def\layerwidth{8*\layerdx}
	\def\layerdepth{7*\layerdy}
	\def\layerheight{\layerdz}

    \foreach \x in {0,2,...,8}
    {
        \draw[layer0] (\x*\layerdx,0,0) -- (\x*\layerdx,\layerdepth,0);
    }
    \foreach \y in {0,1,...,7}
    {
        \draw[layer0] (0,\y*\layerdy,0) -- (\layerwidth,\y*\layerdy,0);
    }
	\foreach \z in {0,1,...,5} {
			\draw[layer\z] (0,0,\layerheight*\z) -- (\layerwidth,0,\layerheight*\z)
			-- (\layerwidth,\layerdepth,\layerheight*\z) -- (0,\layerdepth,\layerheight*\z) -- (0,0,\layerheight*\z) -- cycle;
			\foreach \layernum/\xstart/\ystart/\width/\depth in \highlightboxes
			{
				\ifnum\layernum=\z
					\pgfmathsetmacro{\zcoord}{\layerheight * \layernum}
					\pgfmathsetmacro{\boxheight}{1.0 * \layerdz}

					\coordinate (\ramiprefix-A) at (\xstart*\layerdx, \ystart*\layerdy, \zcoord );
					\coordinate (\ramiprefix-B) at (\xstart*\layerdx + \width*\layerdx, \ystart*\layerdy, \zcoord );
					\coordinate (\ramiprefix-C) at (\xstart*\layerdx + \width*\layerdx, \ystart*\layerdy + \depth*\layerdy, \zcoord );
					\coordinate (\ramiprefix-D) at (\xstart*\layerdx, \ystart*\layerdy + \depth*\layerdy, \zcoord );
					\coordinate (\ramiprefix-E) at (\xstart*\layerdx, \ystart*\layerdy, \zcoord + \boxheight );
					\coordinate (\ramiprefix-F) at (\xstart*\layerdx + \width*\layerdx, \ystart*\layerdy, \zcoord + \boxheight );
					\coordinate (\ramiprefix-G) at (\xstart*\layerdx + \width*\layerdx, \ystart*\layerdy + \depth*\layerdy, \zcoord + \boxheight );
					\coordinate (\ramiprefix-H) at (\xstart*\layerdx, \ystart*\layerdy + \depth*\layerdy, \zcoord + \boxheight );

					\draw[layer\layernum, fill opacity=0.6] (\ramiprefix-A) -- (\ramiprefix-B) -- (\ramiprefix-C) -- (\ramiprefix-D) -- cycle;
					\draw[layer\layernum, fill opacity=0.6] (\ramiprefix-E) -- (\ramiprefix-F) -- (\ramiprefix-G) -- (\ramiprefix-H) -- cycle;
					\draw[layer\layernum, fill opacity=0.6] (\ramiprefix-A) -- (\ramiprefix-B) -- (\ramiprefix-F) -- (\ramiprefix-E) -- cycle;
					\draw[layer\layernum, fill opacity=0.6] (\ramiprefix-B) -- (\ramiprefix-C) -- (\ramiprefix-G) -- (\ramiprefix-F) -- cycle;
					\draw[layer\layernum, fill opacity=0.6] (\ramiprefix-C) -- (\ramiprefix-D) -- (\ramiprefix-H) -- (\ramiprefix-G) -- cycle;
					\draw[layer\layernum, fill opacity=0.6] (\ramiprefix-A) -- (\ramiprefix-D) -- (\ramiprefix-H) -- (\ramiprefix-E) -- cycle;
				\fi
			}
		}

	\foreach \text/\step in {
			Asset/0, Integration/1, Communication/2,
			Information/3, Functional/4, Business/5
		}{
			\coordinate (\ramiprefix-trans-layers-\step) at (0,0,\step*\layerdz + 0.5 * \layerdz);

			\begin{scope}[
					shift=(\ramiprefix-trans-layers-\step),
					rotate around z=0,
					transform shape,
					canvas is xz plane at y=0
				]

				\draw (0,0,0) node[left] {\text};
			\end{scope}
		}

	\foreach \text/\step in {
			Product/0, Field Device/1, Control Device/2,
			Station/3, Work Centers/4, Enterprise/5,
			Connected World/6} {

			\coordinate (\ramiprefix-trans-hierarchy-\step) at (\layerwidth+0.1, \step*\layerdy + 0.5*\layerdy, -0.1);
			\begin{scope}[
					shift=(\ramiprefix-trans-hierarchy-\step),
					transform shape,
					canvas is xz plane at y=0
				]
				\draw (0,0,0) node[right] {\text};
			\end{scope}
		}

	\coordinate (\ramiprefix-trans-lifecycle) at (0, 0, -0.6);
	\begin{scope}[
			shift=(\ramiprefix-trans-lifecycle),
			transform shape,
			canvas is xz plane at y=0
		]
		\small
		\node[draw, fill=shadecolor, align=center,minimum width=2cm, minimum height=3em] at (1,0) {Development};
		\node[draw, fill=shadecolor, align=center,minimum width=2cm, minimum height=3em] at (3,0) {Mainten.\\ Usage};
		\node[draw, fill=shadecolor, align=center,minimum width=2cm, minimum height=3em] at (5,0) {Production};
		\node[draw, fill=shadecolor, align=center,minimum width=2cm, minimum height=3em] at (7,0) {Mainten.\\ Usage};
	\end{scope}

	\coordinate (\ramiprefix-trans-type) at (0, 0, -1.5);
	\begin{scope}[
			shift=(\ramiprefix-trans-type),
			transform shape,
			canvas is xz plane at y=0
		]
		\node[draw, fill=shadecolor, align=center, minimum width=4cm, minimum height=1.7em] at (2,0) {Type};
		\node[draw, fill=shadecolor, align=center, minimum width=4cm, minimum height=1.7em] at (6,0) {Instance};
	\end{scope}

	\coordinate (\ramiprefix-trans-lifecycle-arrow) at (0, 0, 5*\layerdz);
	\begin{scope}[
			shift=(\ramiprefix-trans-lifecycle-arrow),
			transform shape,
			canvas is xz plane at y=0,
		]

		\draw[->] (0,0) -- node[above, align=left]
		{IEC 62890\\ Life Cycle Value Stream}
		++(\layerwidth,0);
	\end{scope}

	\coordinate (\ramiprefix-trans-hierarchy-arrow) at (\layerwidth, 0, 5*\layerdz);
	\begin{scope}[
			shift=(\ramiprefix-trans-hierarchy-arrow),
			transform shape,
			canvas is yz plane at x=0
		]

		\draw[->] (0,0) -- node[above, align=left]
		{IEC 62264, IEC 61512\\ Hierarchy Levels}
		++(\layerdepth,0);
	\end{scope}

	\begin{scope}[
			transform shape,
			canvas is xz plane at y=0
		]

		\draw[->] (0,0) -- node[sloped, below] {Layers} ++(0, 5*\layerdz);
	\end{scope}

\end{tikzpicture}

\ifdefined\includingtikz
\else
\end{document}
\fi
		}
		\caption{The \gls{R4} Model}
		\label{fig:rami}
		\undef\includingtikz
	\end{minipage}%
	\hfill
	\begin{minipage}{0.4\textwidth}
		\centering
		\def\includingtikz{1}
		\scalebox{0.6}{
			\hspace{-2.75cm}
			\ifdefined\includingtikz
\else
   \documentclass{standalone}
   \usepackage{graphicx}
   \usepackage{xcolor}
   \usepackage{tikz}
   \usepackage{pgfplots}
   \pgfplotsset{compat=1.16}
\begin{document}
\fi

\definecolor{maincolor}{RGB}{35, 95, 157}
\definecolor{secondcolor}{RGB}{137, 157, 35}
\definecolor{alertcolor}{RGB}{192, 83, 84}
\definecolor{darkshadecolor}{RGB}{147,112,219}
\colorlet{shadecolor}{darkshadecolor!10}
\colorlet{shadegrey}{black!5}

\def\pyramidwidth{9}    
\def\pyramidheight{6}   

\begin{tikzpicture}[
   layer/.style={darkshadecolor, fill=shadecolor, opacity=0.8}
]
   \draw[layer] (0,0) -- (\pyramidwidth,0) -- (\pyramidwidth/2,\pyramidheight) -- cycle;
   
   \foreach \i [count=\n from 0] in {0,...,4} {
       \pgfmathsetmacro{\ypos}{\i*\pyramidheight/5}
       \pgfmathsetmacro{\width}{\pyramidwidth-(\pyramidwidth*\i/5)}
       \pgfmathsetmacro{\xstart}{(\pyramidwidth-\width)/2}
       \draw[dotted] (\xstart,\ypos) -- (\xstart+\width,\ypos);
   }
   
   \foreach \i [count=\n from 0] in {0,...,4} {
       \pgfmathsetmacro{\ypos}{\i*\pyramidheight/5 + \pyramidheight/10}
       \node[align=center, font=\tiny] at (\pyramidwidth/2, \ypos - 0.23) {
           \ifcase\i
               \textbf{Level 0}\\Field\\Devices
           \or
           \textbf{Level 1}\\Controllers
           \or
           \textbf{Level 2}\\Supervisory\\Control
           \or
           \textbf{Level 3}\\Manufacturing\\Operations
           \or
           \textbf{Level 4}\\Business\\Planning
           \fi
       };
   }
   
   \draw[->] (\pyramidwidth+0.5, 0) -- (\pyramidwidth+0.5, \pyramidheight) 
       node[midway, right, rotate=90,xshift=-89,yshift=5] {\tiny \textbf{Planning Horizon}};
   
   \node[right, font=\tiny, rotate=45] at (\pyramidwidth+0.5, \pyramidheight/5) {Seconds};
   \node[right, font=\tiny, rotate=45] at (\pyramidwidth+0.5, \pyramidheight/2) {Days};
   \node[right, font=\tiny, rotate=45] at (\pyramidwidth+0.5, 4*\pyramidheight/5) {Months};

   \draw[->] (-0.5, 0) -- (-0.5, \pyramidheight)
       node[midway, left, rotate=90,xshift=-35,yshift=-5] {\tiny \textbf{System Types}};
   
   \node[left, font=\tiny, align=right,rotate=45, xshift=-2.5] at (-0.5, \pyramidheight/5) {Sensors,\\Actuators};
   \node[left, font=\tiny, align=right,rotate=45, xshift=-2.5] at (-0.5, 2*\pyramidheight/5) {PLC,\\DCS};
   \node[left, font=\tiny, align=right,rotate=45, xshift=-2.5] at (-0.5, 3*\pyramidheight/5) {SCADA,\\HMI};
   \node[left, font=\tiny, align=right,rotate=45, xshift=-2.5] at (-0.5, 4*\pyramidheight/5) {MES,\\MOM};
   \node[left, font=\tiny, align=right,rotate=45, xshift=-2.5] at (-0.5, 4.8*\pyramidheight/5) {ERP};

\end{tikzpicture}

\ifdefined\includingtikz
\else
\end{document}
\fi
		}
		\caption{The \gls{I95} Model}
		\label{fig:isa95}
		\undef\includingtikz
	\end{minipage}
\end{figure}
\subsubsection{RAMI 4.0.} The \glsfirst{R4} provides a three-dimensional framework that describes the
fundamental aspects of \gls{I4}. It combines business layers, life cycle
stages, and the hierarchy levels of industrial production into a unified model.
The architecture extends traditional automation hierarchies by incorporating
product and work-piece related aspects, thus enabling the representation of
distributed manufacturing systems. \gls{R4}'s layered approach, from asset to
business level, facilitates the integration of existing standards while
providing a foundation for future \gls{I4} implementations.

\cref{fig:rami} shows the \gls{R4} model with these three dimensions: hierarchy
levels, lifecycle stages, and layers. The hierarchy levels extend from the
product level through various manufacturing stages up to the connected world,
reflecting the complete industrial environment. The lifecycle dimension, based
on IEC 62890, represents the product and production lifecycle, distinguishing
between type and instance phases. The layers dimension describes different
aspects from physical assets to business processes, providing a
service-oriented perspective. This layered architecture enables the systematic
integration of various manufacturing components and processes, while
maintaining clear architectural boundaries between functional domains.

\subsubsection{ISA-95} The \gls{I95} standard presents a hierarchical
five-level model for enterprise-control system integration. It defines
standardized terminology and information models that facilitate communication
between enterprise systems and manufacturing operations. This framework enables
consistent system integration while preserving clear functional boundaries. The
hierarchical structure supports organizations in implementing and maintaining
complex automation systems with proper separation of operational
responsibilities.

\cref{fig:isa95} illustrates the hierarchical structure of
\gls{I95}. At Level 4, business planning and logistics handle enterprise-level
activities through \gls{ERP} systems. Level 3 encompasses manufacturing operations
management (MES/MOM), coordinating workflow, recipe management, and production
scheduling. Level 2 consists of supervisory control systems (\gls{SCADA}) for
monitoring and controlling multiple processes. Level 1 contains the direct
control systems (PLCs, DCS) that manage specific production processes, while
Level 0 comprises the actual production process with its sensors and actuators.
Each level operates on different time scales, from months at Level 4 down to
seconds or milliseconds at Level 0, reflecting the distinct operational
requirements at each layer of the automation hierarchy.

\section{Challenges of MLOps in OT}\label{sec:challenges}
Combining \gls{MLOps} with \gls{OT} systems requires a thorough understanding
of each domain's guiding principles and constraints. While \gls{MLOps} has
found its use in IT-centric applications, \gls{OT} is still lacking
adoption because it demands highly reliable and real-time processes.
Particularly, \gls{OT} is governed by safety and compatibility standards like
NIST 800-82~\cite{nist80082}, which provides guidance on securing \gls{OT}
systems while addressing their unique requirements \cite{Faubel2023MLOps4.0}.
Additionally frameworks like \gls{I95} and \gls{R4} structure data,
control, and lifecycles in a different way than solely software based \gls{IT}
environments. The following sections examine the key challenges and
characteristics that emerge when integrating \gls{MLOps} into \gls{OT}
environments.

\subsection{Core OT System Requirements}

\subsubsection{Real-time and Deterministic Requirements.} Many \gls{OT} systems have
real-time operations as a core task. \gls{OT} demands strict timing
constraints, which NIST 800-82~\cite{nist80082} additionally emphasizes. In practice, slight
latency or jitter can lead to suboptimal operation which can damage machines or
endanger humans. This creates significant challenges for ML model deployment
and inference, requiring specialized frameworks and careful consideration of
processing locations to meet timing requirements~\cite{mazlan2024review}.

\subsubsection{High Reliability and Availability.} \gls{OT} systems cannot tolerate
unplanned downtime as it leads to safety incidents, damaged equipment, and lost
productivity. These high stakes require redundancy and fault-tolerant
architectures by design~\cite{mazlan2024review}. Common \gls{MLOps} practices like rolling updates or
iterative testing become problematic when dealing with real-time \gls{OT} systems
running 24/7, necessitating new approaches to model deployment and updates.

\subsection{Integration and Infrastructure Challenges}

\subsubsection{Infrastructure and Resource Constraints.} Edge devices and
industrial systems have limited computational capabilities compared to
traditional IT infrastructure. This requires specialized deployment automation
and frameworks, often necessitating lightweight versions of ML frameworks
designed specifically for edge deployment. The heterogeneous nature of \gls{OT}
hardware, communication protocols and operating systems further complicates
testing and deployment across diverse legacy equipment and modern automation
systems.

\subsubsection{Data and Integration Requirements.} Industrial systems demand
real-time processing and low-latency decision making capabilities.
Mission-critical data often cannot leave on-premises due to volume or security
concerns, requiring edge processing solutions. Systems must integrate with
existing \gls{MES} and \gls{ERP} systems while maintaining bi-directional
synchronization between physical systems and their digital twins.

\subsection{Operational and Compliance Considerations}

\subsubsection{Safety-Critical and Regulated Environments.} NIST
800.82 emphasizes cybersecurity importance, as breaches or
system malfunctions can lead to physical hazards. \gls{OT} settings must comply
with domain-specific regulations like IEC 62443 for industrial
cybersecurity or ISO standards for quality management. This compliance
landscape creates strict constraints on how new \gls{ml} models can be
integrated and validated.

\subsubsection{Validation and Testing Constraints.} The industrial context
severely limits in-production testing practices common in IT \gls{MLOps}.
Organizations must develop specialized isolated sandboxes that accurately
simulate industrial environments for testing. Strict waterfall-like engineering
standards in industrial settings limit continuous deployment practices,
requiring careful adaptation of \gls{MLOps} principles to work within existing
processes.

\subsubsection{Architectural Considerations.} \gls{OT} systems vary
significantly in their architectural approach. Some rely on centralized \gls{SCADA}
systems, while others distribute control logic to edge devices or local PLC
networks. ML-based solutions might run on edge devices on the shop floor to
bypass cloud dependencies. As NIST 800-82~\cite{nist80082} documents, architectures can range
from deeply hierarchical setups to more modern, service-oriented ones, making
it clear that standard cloud-based \gls{MLOps} approaches rarely apply unmodified in
\gls{OT} environments.

\section{Embedding MLOps into OT Reference Models}

\subsection{ISA-95 and MLOps}
The \gls{I95} standard is a comprehensive framework, but because of its
comprehensive nature and focus on enterprise-control system integration, it
lacks the granularity in the hierarchy dimension as well as further dimensions to
provide a detailed mapping of \gls{MLOps} components within an industrial
production environment. When mapping \gls{MLOps} lifecycle phases to \gls{I95},
many of the phases can be located in multiple, if not all, levels of the
model. For example, \textbf{Model Deployment} can occur at the lowest level on
the shop floor, but also at the enterprise level, depending on the use case.
Similarly, \textbf{Continuous Training} can be performed at the lowest level,
but also at the enterprise level, depending on the data sources and training
architecture. This ambiguity in placement makes it challenging to establish
clear governance boundaries and responsibilities.

Since this provides no additional insights into the integration of \gls{MLOps}
into \gls{OT} systems, we restrict our focus to the \gls{R4} model, which
provides more structure than \gls{I95} by incorporating further dimensions and
also adds granularity in the hierarchy dimension. The \gls{R4} framework's
three-dimensional approach (hierarchy levels, lifecycle value stream, and
layers) enables a more precise mapping of \gls{MLOps} components within
industrial environments, facilitating better alignment between data science
workflows and \gls{OT} infrastructure.

\subsection{RAMI 4.0 and MLOps}

To ensure the dynamic applicability of \gls{R4} and provide a structured
framework for the industrial application of \gls{MLOps}, a systematic mapping
is required. This mapping aligns \gls{MLOps} phases with \gls{R4}'s hierarchy
levels, lifecycle stages, and architectural layers. By structuring this
mapping, organizations can bridge the gap between \gls{ml} workflows
and industrial automation processes. It facilitates interoperability,
traceability, and efficient system integration while ensuring that \gls{MLOps}
remains adaptable to evolving industry requirements. Table \ref{table:mapping}
outlines the mapping, which is subsequently summarized and explained. Finally,
the mapping is evaluated within a real-world use case.

\begin{table}[t] \centering \caption{\gls{MLOps} Mapped to \gls{R4}} \label{table:mapping}
	\begin{tabular}{p{3cm} p{6cm} p{3cm}} \toprule \textbf{\gls{MLOps} Phase} & \textbf{\gls{R4} Dimensions}                                                                                   & \textbf{Tasks}                                         \\ \midrule
               Model\newline Development                                  & Hierarchy: Station/Enterprise\newline Lifecycle: Type/Development\newline Layers: Communication \& Information & Development of \gls{ml} models                         \\ \midrule
               Training \newline Operationalization                       & Hierarchy: Station/Enterprise\newline Lifecycle: Type/Usage\newline Layers: Communication \& Information       & Data Preprocessing \& Development of training pipeline \\ \midrule
               Continuous \newline Training                               & Hierarchy: Station/Enterprise\newline Lifecycle: Instance/Usage\newline Layers: Communication \& Information   & (Re-)Training of models with training pipeline         \\ \midrule
               Model \newline Deployment                                  & Hierarchy: Field/Control/Enterprise\newline Lifecycle: Type/Usage\newline Layers: Function                     & Deployment models on inference servers or edge devices \\ \midrule
               Prediction \newline Serving                                & Hierarchy: Field--Enterprise\newline Lifecycle: Instance/Production\newline Layers: Asset                      & Application models during production                   \\ \midrule
               Continuous \newline Monitoring                             & Hierarchy: Field--Enterprise\newline Lifecycle: Instance/Usage\newline Layers: Function \& Business            & Monitoring of models and data \& Triggering of updates \\ \bottomrule
	\end{tabular}
\end{table}

The following elaborates on the summary in \cref{table:mapping} in more detail:
\subsubsection{Hierarchy Levels}
\begin{itemize}
	\item \textbf{Product/Field Device/Control Device}: Sensors and
	      actuators collect data for training and inference. When
	      latency requirements are strict, inference can be executed on
	      embedded systems at the Field Device and Control Device
	      levels. This directly addresses the "Real-time and
	      Deterministic Requirements" challenge by minimizing network
	      latency and ensuring time-critical processing occurs close to
	      the physical process.
	\item                  \textbf{Station/Enterprise/Connected World}: Model training,
	      experiment tracking, and orchestration occur at higher
	      hierarchy levels, leveraging cloud platforms or on-premises high
	      performance computing clusters for complex training jobs.
	      This approach mitigates the "Infrastructure and Resource
	      Constraints" challenge by placing computationally intensive
	      operations where sufficient resources exist while maintaining
	      operational separation from critical control systems.
\end{itemize}
\subsubsection{Lifecycle and Value Stream}
\begin{itemize}
	\item \textbf{Development \& Production}: The core \gls{ml} phases
	      (data ingestion, training, deployment, monitoring) are embedded
	      into the standard lifecycle phases of product and system
	      engineering. During Type phases (system design), data
	      requirements and model training strategies are established. In
	      Instance phases (production runtime), continuously trained models are
	      deployed. This structured approach addresses the
	      \enquote{Validation and Testing Constraints} challenge by providing
	      clear separation between development and production
	      environments while maintaining compliance with industrial
	      engineering standards.
	\item \textbf{Operation \& Maintenance}: \gls{R4} maintenance
	      phases naturally align with continuous retraining and model
	      updates in \gls{MLOps}. Performance monitoring in production
	      helps identify optimization potential, leading to triggered
	      deployment of improved models. This alignment addresses the
	      \enquote{High Reliability and Availability} challenge by
	      integrating model updates into existing maintenance
	      frameworks that preserve system stability and operational
	      continuity.
\end{itemize}
\subsubsection{Layers}
\begin{itemize}
	\item \textbf{Asset \& Integration Layers}: Physical sensors feed
	      raw signals to integration components adding higher level
	      context about the production process and unify heterogeneous
	      protocols (fieldbuses, industrial Ethernet) into a normalized
	      data pipeline, located within \gls{MLOps} data management.
	      Using \gls{OPCUA} as a communication protocol, this
	      integration addresses the \enquote{Data and Integration Requirements}
	      challenge by enabling the use of specialized components for
	      \gls{ml} as well as legacy equipment. In the inference phase
	      Assets in all hierarchy levels can be actors executing model
	      decisions.
	\item \textbf{Communication \& Information Layers}: These layers
	      securely host the training data, model parameters and
	      source code. Metadata about experiments, training runs, and
	      performance are captured to ensure interoperability and
	      traceability. This approach addresses the "Safety-Critical
	      and Regulated Environments" challenge by providing
	      comprehensive documentation and auditability required for
	      regulatory compliance.
	\item \textbf{Functional \& Business Layers}: Model
	      orchestration, decision-support logic, and management
	      dashboards align with the Functional Layer. The Business
	      Layer focuses on enterprise-level objectives, such as
	      reducing downtime or improving throughput. This alignment
	      enables data-driven decision making at appropriate
	      organizational levels.
\end{itemize}

Data and Model Management is done in parallel to the other six phases, as it is
centered in the \gls{MLOps} cycle. Thus, it is also not placed in \gls{R4} but rather
as a parallel task next to the three-dimensional cube influencing all layers,
hierarchy levels, or lifecycle phases. To give a more detailed specification,
data management is mainly performed on the Asset and Integration Layers, as it
originates from real production system components. Model management, on the
other hand, is primarily handled in the Communication and Information Layers,
where model metadata, versioning, and interoperability are ensured.

All these alignments address the \enquote{Architectural Considerations} challenge by
respecting the existing operational structures while enhancing them with
data-driven decision-making capabilities.
\newcounter{ramiCounter}

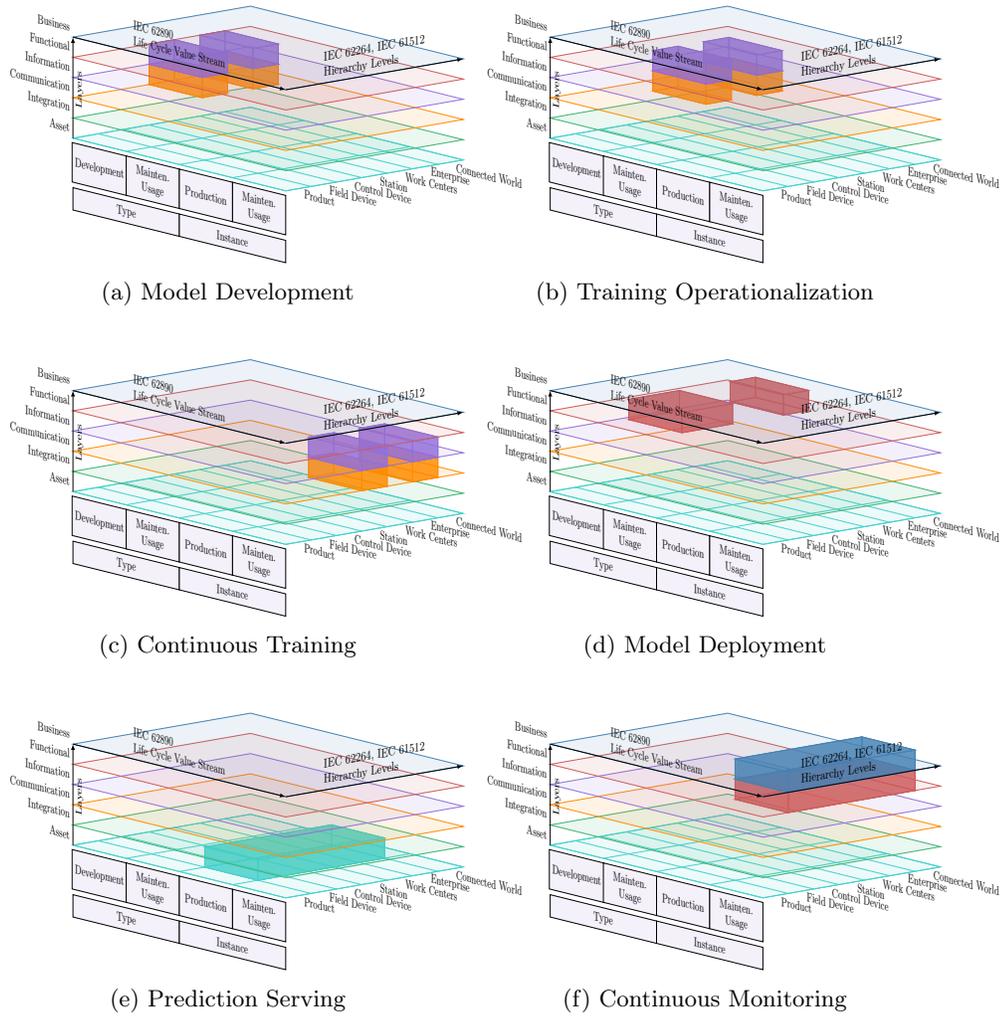
\begin{figure}[htbp]
	\centering

	\begin{minipage}{0.48\textwidth}
		\centering
		\def\includingtikz{1}
		\def\ramiprefix{A1}
		\def\highlightboxes{2/0/3/2/1,3/0/3/2/1,2/0/5/2/1,3/0/5/2/1}
		\scalebox{0.55}{
			\hspace{-2.5cm}
			\ifdefined\includingtikz
    \ifdefined\highlightboxes
    \else
        \def\highlightboxes{}
    \fi
\else
    \documentclass{standalone}
    \usepackage{graphicx}
    \usepackage{xcolor}
    \usepackage{tikz}
    \usepackage{tikz-3dplot}
    \usepackage{pgfplots}
    \pgfplotsset{compat=1.16}
    \def\highlightboxes{
			}
\begin{document}
\fi

\ifdefined\ramiprefix
\else
	\def\ramiprefix{X}
\fi

\definecolor{maincolor}{RGB}{35, 95, 157}
\definecolor{secondcolor}{RGB}{137, 157, 35}
\definecolor{alertcolor}{RGB}{192, 83, 84}
\definecolor{darkshadecolor}{RGB}{147,112,219}
\colorlet{shadecolor}{darkshadecolor!10}
\colorlet{shadegrey}{black!5}
\definecolor{cLayer1}{RGB}{72,209,204}
\definecolor{cLayer2}{RGB}{60,179,113}
\definecolor{cLayer3}{RGB}{255,140,0}
\definecolor{cLayer4}{RGB}{147,112,219}
\definecolor{cLayer5}{RGB}{205,92,92}
\definecolor{cLayer6}{RGB}{70,130,180}

\usetikzlibrary{calc, shapes.symbols, shapes.misc, shapes.arrows, decorations,
	decorations.pathmorphing, fit, matrix, matrix.skeleton, backgrounds}
\tikzset{>=latex}

\tikzstyle{boxnodewhite}=[draw, rectangle, minimum height=2em]
\tikzstyle{boxnode}=[boxnodewhite, fill=shadecolor]
\tikzstyle{boxnodehalf}=[boxnode, dashed, opacity=0.5]

\makeatletter
\newdimen\multi@col@width
\newdimen\multi@col@margin
\newcount\multi@col@count
\multi@col@width=0pt

\tikzset{
	multicol/.code={%
			\global\multi@col@count=#1\relax
			\global\let\orig@pgfmatrixendcode=\pgfmatrixendcode
			\global\let\orig@pgfmatrixemptycode=\pgfmatrixemptycode
			\def\pgfmatrixendcode##1{\orig@pgfmatrixendcode%
				##1%
				\pgfutil@tempdima=\pgf@picmaxx
				\global\multi@col@margin=\pgf@picminx
				\advance\pgfutil@tempdima by -\pgf@picminx
				\divide\pgfutil@tempdima by #1\relax
				\global\multi@col@width=\pgfutil@tempdima
				\pgf@picmaxx=.5\multi@col@width
				\pgf@picminx=-.5\multi@col@width
				\global\pgf@picmaxx=\pgf@picmaxx
				\global\pgf@picminx=\pgf@picminx
				\gdef\multi@adjust@position{%
					\setbox\pgf@matrix@cell=\hbox\bgroup
					\hfil\hskip-1.5\multi@col@margin
					\hfil\hskip-.5\multi@col@width
					\box\pgf@matrix@cell
					\egroup
				}%
				\gdef\multi@temp{\aftergroup\multi@adjust@position}%
				\aftergroup\multi@temp
			}
			\gdef\pgfmatrixemptycode{%
				\orig@pgfmatrixemptycode
				\global\advance\multi@col@count by -1\relax
				\global\pgf@picmaxx=.5\multi@col@width
				\global\pgf@picminx=-.5\multi@col@width
				\ifnum\multi@col@count=1\relax
					\global\let\pgfmatrixemptycode=\orig@pgfmatrixemptycode
				\fi
			}
		}
}
\makeatother

\tdplotsetmaincoords{78}{50}
\begin{tikzpicture}[
		tdplot_main_coords,
		layer/.style={draw opacity=1.0, fill opacity=0.1},
		layer0/.style={layer, draw=cLayer1, fill=cLayer1},
		layer1/.style={layer, draw=cLayer2, fill=cLayer2},
		layer2/.style={layer, draw=cLayer3, fill=cLayer3},
		layer3/.style={layer, draw=cLayer4, fill=cLayer4},
		layer4/.style={layer, draw=cLayer5, fill=cLayer5},
		layer5/.style={layer, draw=cLayer6, fill=cLayer6}
	]
	\def\layerdy{0.8}
	\def\layerdz{0.5}
    \def\layerdx{1}

	\def\layerwidth{8*\layerdx}
	\def\layerdepth{7*\layerdy}
	\def\layerheight{\layerdz}

    \foreach \x in {0,2,...,8}
    {
        \draw[layer0] (\x*\layerdx,0,0) -- (\x*\layerdx,\layerdepth,0);
    }
    \foreach \y in {0,1,...,7}
    {
        \draw[layer0] (0,\y*\layerdy,0) -- (\layerwidth,\y*\layerdy,0);
    }
	\foreach \z in {0,1,...,5} {
			\draw[layer\z] (0,0,\layerheight*\z) -- (\layerwidth,0,\layerheight*\z)
			-- (\layerwidth,\layerdepth,\layerheight*\z) -- (0,\layerdepth,\layerheight*\z) -- (0,0,\layerheight*\z) -- cycle;
			\foreach \layernum/\xstart/\ystart/\width/\depth in \highlightboxes
			{
				\ifnum\layernum=\z
					\pgfmathsetmacro{\zcoord}{\layerheight * \layernum}
					\pgfmathsetmacro{\boxheight}{1.0 * \layerdz}

					\coordinate (\ramiprefix-A) at (\xstart*\layerdx, \ystart*\layerdy, \zcoord );
					\coordinate (\ramiprefix-B) at (\xstart*\layerdx + \width*\layerdx, \ystart*\layerdy, \zcoord );
					\coordinate (\ramiprefix-C) at (\xstart*\layerdx + \width*\layerdx, \ystart*\layerdy + \depth*\layerdy, \zcoord );
					\coordinate (\ramiprefix-D) at (\xstart*\layerdx, \ystart*\layerdy + \depth*\layerdy, \zcoord );
					\coordinate (\ramiprefix-E) at (\xstart*\layerdx, \ystart*\layerdy, \zcoord + \boxheight );
					\coordinate (\ramiprefix-F) at (\xstart*\layerdx + \width*\layerdx, \ystart*\layerdy, \zcoord + \boxheight );
					\coordinate (\ramiprefix-G) at (\xstart*\layerdx + \width*\layerdx, \ystart*\layerdy + \depth*\layerdy, \zcoord + \boxheight );
					\coordinate (\ramiprefix-H) at (\xstart*\layerdx, \ystart*\layerdy + \depth*\layerdy, \zcoord + \boxheight );

					\draw[layer\layernum, fill opacity=0.6] (\ramiprefix-A) -- (\ramiprefix-B) -- (\ramiprefix-C) -- (\ramiprefix-D) -- cycle;
					\draw[layer\layernum, fill opacity=0.6] (\ramiprefix-E) -- (\ramiprefix-F) -- (\ramiprefix-G) -- (\ramiprefix-H) -- cycle;
					\draw[layer\layernum, fill opacity=0.6] (\ramiprefix-A) -- (\ramiprefix-B) -- (\ramiprefix-F) -- (\ramiprefix-E) -- cycle;
					\draw[layer\layernum, fill opacity=0.6] (\ramiprefix-B) -- (\ramiprefix-C) -- (\ramiprefix-G) -- (\ramiprefix-F) -- cycle;
					\draw[layer\layernum, fill opacity=0.6] (\ramiprefix-C) -- (\ramiprefix-D) -- (\ramiprefix-H) -- (\ramiprefix-G) -- cycle;
					\draw[layer\layernum, fill opacity=0.6] (\ramiprefix-A) -- (\ramiprefix-D) -- (\ramiprefix-H) -- (\ramiprefix-E) -- cycle;
				\fi
			}
		}

	\foreach \text/\step in {
			Asset/0, Integration/1, Communication/2,
			Information/3, Functional/4, Business/5
		}{
			\coordinate (\ramiprefix-trans-layers-\step) at (0,0,\step*\layerdz + 0.5 * \layerdz);

			\begin{scope}[
					shift=(\ramiprefix-trans-layers-\step),
					rotate around z=0,
					transform shape,
					canvas is xz plane at y=0
				]

				\draw (0,0,0) node[left] {\text};
			\end{scope}
		}

	\foreach \text/\step in {
			Product/0, Field Device/1, Control Device/2,
			Station/3, Work Centers/4, Enterprise/5,
			Connected World/6} {

			\coordinate (\ramiprefix-trans-hierarchy-\step) at (\layerwidth+0.1, \step*\layerdy + 0.5*\layerdy, -0.1);
			\begin{scope}[
					shift=(\ramiprefix-trans-hierarchy-\step),
					transform shape,
					canvas is xz plane at y=0
				]
				\draw (0,0,0) node[right] {\text};
			\end{scope}
		}

	\coordinate (\ramiprefix-trans-lifecycle) at (0, 0, -0.6);
	\begin{scope}[
			shift=(\ramiprefix-trans-lifecycle),
			transform shape,
			canvas is xz plane at y=0
		]
		\small
		\node[draw, fill=shadecolor, align=center,minimum width=2cm, minimum height=3em] at (1,0) {Development};
		\node[draw, fill=shadecolor, align=center,minimum width=2cm, minimum height=3em] at (3,0) {Mainten.\\ Usage};
		\node[draw, fill=shadecolor, align=center,minimum width=2cm, minimum height=3em] at (5,0) {Production};
		\node[draw, fill=shadecolor, align=center,minimum width=2cm, minimum height=3em] at (7,0) {Mainten.\\ Usage};
	\end{scope}

	\coordinate (\ramiprefix-trans-type) at (0, 0, -1.5);
	\begin{scope}[
			shift=(\ramiprefix-trans-type),
			transform shape,
			canvas is xz plane at y=0
		]
		\node[draw, fill=shadecolor, align=center, minimum width=4cm, minimum height=1.7em] at (2,0) {Type};
		\node[draw, fill=shadecolor, align=center, minimum width=4cm, minimum height=1.7em] at (6,0) {Instance};
	\end{scope}

	\coordinate (\ramiprefix-trans-lifecycle-arrow) at (0, 0, 5*\layerdz);
	\begin{scope}[
			shift=(\ramiprefix-trans-lifecycle-arrow),
			transform shape,
			canvas is xz plane at y=0,
		]

		\draw[->] (0,0) -- node[above, align=left]
		{IEC 62890\\ Life Cycle Value Stream}
		++(\layerwidth,0);
	\end{scope}

	\coordinate (\ramiprefix-trans-hierarchy-arrow) at (\layerwidth, 0, 5*\layerdz);
	\begin{scope}[
			shift=(\ramiprefix-trans-hierarchy-arrow),
			transform shape,
			canvas is yz plane at x=0
		]

		\draw[->] (0,0) -- node[above, align=left]
		{IEC 62264, IEC 61512\\ Hierarchy Levels}
		++(\layerdepth,0);
	\end{scope}

	\begin{scope}[
			transform shape,
			canvas is xz plane at y=0
		]

		\draw[->] (0,0) -- node[sloped, below] {Layers} ++(0, 5*\layerdz);
	\end{scope}

\end{tikzpicture}

\ifdefined\includingtikz
\else
\end{document}
\fi
		}
		\subcaption{Model Development}
		\label{fig:rami1}
		\undef\includingtikz
		\undef\ramiprefix
		\undef\highlightboxes
	\end{minipage}%
	\hfill
	\begin{minipage}{0.48\textwidth}
		\centering
		\def\includingtikz{1}
		\def\ramiprefix{A2}
		\def\highlightboxes{2/1/3/2/1,3/1/3/2/1,2/1/5/2/1,3/1/5/2/1}
		\scalebox{0.55}{
			\hspace{-2.5cm}
			\ifdefined\includingtikz
    \ifdefined\highlightboxes
    \else
        \def\highlightboxes{}
    \fi
\else
    \documentclass{standalone}
    \usepackage{graphicx}
    \usepackage{xcolor}
    \usepackage{tikz}
    \usepackage{tikz-3dplot}
    \usepackage{pgfplots}
    \pgfplotsset{compat=1.16}
    \def\highlightboxes{
			}
\begin{document}
\fi

\ifdefined\ramiprefix
\else
	\def\ramiprefix{X}
\fi

\definecolor{maincolor}{RGB}{35, 95, 157}
\definecolor{secondcolor}{RGB}{137, 157, 35}
\definecolor{alertcolor}{RGB}{192, 83, 84}
\definecolor{darkshadecolor}{RGB}{147,112,219}
\colorlet{shadecolor}{darkshadecolor!10}
\colorlet{shadegrey}{black!5}
\definecolor{cLayer1}{RGB}{72,209,204}
\definecolor{cLayer2}{RGB}{60,179,113}
\definecolor{cLayer3}{RGB}{255,140,0}
\definecolor{cLayer4}{RGB}{147,112,219}
\definecolor{cLayer5}{RGB}{205,92,92}
\definecolor{cLayer6}{RGB}{70,130,180}

\usetikzlibrary{calc, shapes.symbols, shapes.misc, shapes.arrows, decorations,
	decorations.pathmorphing, fit, matrix, matrix.skeleton, backgrounds}
\tikzset{>=latex}

\tikzstyle{boxnodewhite}=[draw, rectangle, minimum height=2em]
\tikzstyle{boxnode}=[boxnodewhite, fill=shadecolor]
\tikzstyle{boxnodehalf}=[boxnode, dashed, opacity=0.5]

\makeatletter
\newdimen\multi@col@width
\newdimen\multi@col@margin
\newcount\multi@col@count
\multi@col@width=0pt

\tikzset{
	multicol/.code={%
			\global\multi@col@count=#1\relax
			\global\let\orig@pgfmatrixendcode=\pgfmatrixendcode
			\global\let\orig@pgfmatrixemptycode=\pgfmatrixemptycode
			\def\pgfmatrixendcode##1{\orig@pgfmatrixendcode%
				##1%
				\pgfutil@tempdima=\pgf@picmaxx
				\global\multi@col@margin=\pgf@picminx
				\advance\pgfutil@tempdima by -\pgf@picminx
				\divide\pgfutil@tempdima by #1\relax
				\global\multi@col@width=\pgfutil@tempdima
				\pgf@picmaxx=.5\multi@col@width
				\pgf@picminx=-.5\multi@col@width
				\global\pgf@picmaxx=\pgf@picmaxx
				\global\pgf@picminx=\pgf@picminx
				\gdef\multi@adjust@position{%
					\setbox\pgf@matrix@cell=\hbox\bgroup
					\hfil\hskip-1.5\multi@col@margin
					\hfil\hskip-.5\multi@col@width
					\box\pgf@matrix@cell
					\egroup
				}%
				\gdef\multi@temp{\aftergroup\multi@adjust@position}%
				\aftergroup\multi@temp
			}
			\gdef\pgfmatrixemptycode{%
				\orig@pgfmatrixemptycode
				\global\advance\multi@col@count by -1\relax
				\global\pgf@picmaxx=.5\multi@col@width
				\global\pgf@picminx=-.5\multi@col@width
				\ifnum\multi@col@count=1\relax
					\global\let\pgfmatrixemptycode=\orig@pgfmatrixemptycode
				\fi
			}
		}
}
\makeatother

\tdplotsetmaincoords{78}{50}
\begin{tikzpicture}[
		tdplot_main_coords,
		layer/.style={draw opacity=1.0, fill opacity=0.1},
		layer0/.style={layer, draw=cLayer1, fill=cLayer1},
		layer1/.style={layer, draw=cLayer2, fill=cLayer2},
		layer2/.style={layer, draw=cLayer3, fill=cLayer3},
		layer3/.style={layer, draw=cLayer4, fill=cLayer4},
		layer4/.style={layer, draw=cLayer5, fill=cLayer5},
		layer5/.style={layer, draw=cLayer6, fill=cLayer6}
	]
	\def\layerdy{0.8}
	\def\layerdz{0.5}
    \def\layerdx{1}

	\def\layerwidth{8*\layerdx}
	\def\layerdepth{7*\layerdy}
	\def\layerheight{\layerdz}

    \foreach \x in {0,2,...,8}
    {
        \draw[layer0] (\x*\layerdx,0,0) -- (\x*\layerdx,\layerdepth,0);
    }
    \foreach \y in {0,1,...,7}
    {
        \draw[layer0] (0,\y*\layerdy,0) -- (\layerwidth,\y*\layerdy,0);
    }
	\foreach \z in {0,1,...,5} {
			\draw[layer\z] (0,0,\layerheight*\z) -- (\layerwidth,0,\layerheight*\z)
			-- (\layerwidth,\layerdepth,\layerheight*\z) -- (0,\layerdepth,\layerheight*\z) -- (0,0,\layerheight*\z) -- cycle;
			\foreach \layernum/\xstart/\ystart/\width/\depth in \highlightboxes
			{
				\ifnum\layernum=\z
					\pgfmathsetmacro{\zcoord}{\layerheight * \layernum}
					\pgfmathsetmacro{\boxheight}{1.0 * \layerdz}

					\coordinate (\ramiprefix-A) at (\xstart*\layerdx, \ystart*\layerdy, \zcoord );
					\coordinate (\ramiprefix-B) at (\xstart*\layerdx + \width*\layerdx, \ystart*\layerdy, \zcoord );
					\coordinate (\ramiprefix-C) at (\xstart*\layerdx + \width*\layerdx, \ystart*\layerdy + \depth*\layerdy, \zcoord );
					\coordinate (\ramiprefix-D) at (\xstart*\layerdx, \ystart*\layerdy + \depth*\layerdy, \zcoord );
					\coordinate (\ramiprefix-E) at (\xstart*\layerdx, \ystart*\layerdy, \zcoord + \boxheight );
					\coordinate (\ramiprefix-F) at (\xstart*\layerdx + \width*\layerdx, \ystart*\layerdy, \zcoord + \boxheight );
					\coordinate (\ramiprefix-G) at (\xstart*\layerdx + \width*\layerdx, \ystart*\layerdy + \depth*\layerdy, \zcoord + \boxheight );
					\coordinate (\ramiprefix-H) at (\xstart*\layerdx, \ystart*\layerdy + \depth*\layerdy, \zcoord + \boxheight );

					\draw[layer\layernum, fill opacity=0.6] (\ramiprefix-A) -- (\ramiprefix-B) -- (\ramiprefix-C) -- (\ramiprefix-D) -- cycle;
					\draw[layer\layernum, fill opacity=0.6] (\ramiprefix-E) -- (\ramiprefix-F) -- (\ramiprefix-G) -- (\ramiprefix-H) -- cycle;
					\draw[layer\layernum, fill opacity=0.6] (\ramiprefix-A) -- (\ramiprefix-B) -- (\ramiprefix-F) -- (\ramiprefix-E) -- cycle;
					\draw[layer\layernum, fill opacity=0.6] (\ramiprefix-B) -- (\ramiprefix-C) -- (\ramiprefix-G) -- (\ramiprefix-F) -- cycle;
					\draw[layer\layernum, fill opacity=0.6] (\ramiprefix-C) -- (\ramiprefix-D) -- (\ramiprefix-H) -- (\ramiprefix-G) -- cycle;
					\draw[layer\layernum, fill opacity=0.6] (\ramiprefix-A) -- (\ramiprefix-D) -- (\ramiprefix-H) -- (\ramiprefix-E) -- cycle;
				\fi
			}
		}

	\foreach \text/\step in {
			Asset/0, Integration/1, Communication/2,
			Information/3, Functional/4, Business/5
		}{
			\coordinate (\ramiprefix-trans-layers-\step) at (0,0,\step*\layerdz + 0.5 * \layerdz);

			\begin{scope}[
					shift=(\ramiprefix-trans-layers-\step),
					rotate around z=0,
					transform shape,
					canvas is xz plane at y=0
				]

				\draw (0,0,0) node[left] {\text};
			\end{scope}
		}

	\foreach \text/\step in {
			Product/0, Field Device/1, Control Device/2,
			Station/3, Work Centers/4, Enterprise/5,
			Connected World/6} {

			\coordinate (\ramiprefix-trans-hierarchy-\step) at (\layerwidth+0.1, \step*\layerdy + 0.5*\layerdy, -0.1);
			\begin{scope}[
					shift=(\ramiprefix-trans-hierarchy-\step),
					transform shape,
					canvas is xz plane at y=0
				]
				\draw (0,0,0) node[right] {\text};
			\end{scope}
		}

	\coordinate (\ramiprefix-trans-lifecycle) at (0, 0, -0.6);
	\begin{scope}[
			shift=(\ramiprefix-trans-lifecycle),
			transform shape,
			canvas is xz plane at y=0
		]
		\small
		\node[draw, fill=shadecolor, align=center,minimum width=2cm, minimum height=3em] at (1,0) {Development};
		\node[draw, fill=shadecolor, align=center,minimum width=2cm, minimum height=3em] at (3,0) {Mainten.\\ Usage};
		\node[draw, fill=shadecolor, align=center,minimum width=2cm, minimum height=3em] at (5,0) {Production};
		\node[draw, fill=shadecolor, align=center,minimum width=2cm, minimum height=3em] at (7,0) {Mainten.\\ Usage};
	\end{scope}

	\coordinate (\ramiprefix-trans-type) at (0, 0, -1.5);
	\begin{scope}[
			shift=(\ramiprefix-trans-type),
			transform shape,
			canvas is xz plane at y=0
		]
		\node[draw, fill=shadecolor, align=center, minimum width=4cm, minimum height=1.7em] at (2,0) {Type};
		\node[draw, fill=shadecolor, align=center, minimum width=4cm, minimum height=1.7em] at (6,0) {Instance};
	\end{scope}

	\coordinate (\ramiprefix-trans-lifecycle-arrow) at (0, 0, 5*\layerdz);
	\begin{scope}[
			shift=(\ramiprefix-trans-lifecycle-arrow),
			transform shape,
			canvas is xz plane at y=0,
		]

		\draw[->] (0,0) -- node[above, align=left]
		{IEC 62890\\ Life Cycle Value Stream}
		++(\layerwidth,0);
	\end{scope}

	\coordinate (\ramiprefix-trans-hierarchy-arrow) at (\layerwidth, 0, 5*\layerdz);
	\begin{scope}[
			shift=(\ramiprefix-trans-hierarchy-arrow),
			transform shape,
			canvas is yz plane at x=0
		]

		\draw[->] (0,0) -- node[above, align=left]
		{IEC 62264, IEC 61512\\ Hierarchy Levels}
		++(\layerdepth,0);
	\end{scope}

	\begin{scope}[
			transform shape,
			canvas is xz plane at y=0
		]

		\draw[->] (0,0) -- node[sloped, below] {Layers} ++(0, 5*\layerdz);
	\end{scope}

\end{tikzpicture}

\ifdefined\includingtikz
\else
\end{document}
\fi
		}
		\subcaption{Training Operationalization}
		\label{fig:rami2}
		\undef\includingtikz
		\undef\ramiprefix
		\undef\highlightboxes
	\end{minipage}

	\vspace{2em}

	\begin{minipage}{0.48\textwidth}
		\centering
		\def\includingtikz{1}
		\def\ramiprefix{A3}
		\def\highlightboxes{2/6/3/2/1,3/6/3/2/1,2/6/5/2/1,3/6/5/2/1}

		\scalebox{0.55}{
			\hspace{-2.5cm}
			\ifdefined\includingtikz
    \ifdefined\highlightboxes
    \else
        \def\highlightboxes{}
    \fi
\else
    \documentclass{standalone}
    \usepackage{graphicx}
    \usepackage{xcolor}
    \usepackage{tikz}
    \usepackage{tikz-3dplot}
    \usepackage{pgfplots}
    \pgfplotsset{compat=1.16}
    \def\highlightboxes{
			}
\begin{document}
\fi

\ifdefined\ramiprefix
\else
	\def\ramiprefix{X}
\fi

\definecolor{maincolor}{RGB}{35, 95, 157}
\definecolor{secondcolor}{RGB}{137, 157, 35}
\definecolor{alertcolor}{RGB}{192, 83, 84}
\definecolor{darkshadecolor}{RGB}{147,112,219}
\colorlet{shadecolor}{darkshadecolor!10}
\colorlet{shadegrey}{black!5}
\definecolor{cLayer1}{RGB}{72,209,204}
\definecolor{cLayer2}{RGB}{60,179,113}
\definecolor{cLayer3}{RGB}{255,140,0}
\definecolor{cLayer4}{RGB}{147,112,219}
\definecolor{cLayer5}{RGB}{205,92,92}
\definecolor{cLayer6}{RGB}{70,130,180}

\usetikzlibrary{calc, shapes.symbols, shapes.misc, shapes.arrows, decorations,
	decorations.pathmorphing, fit, matrix, matrix.skeleton, backgrounds}
\tikzset{>=latex}

\tikzstyle{boxnodewhite}=[draw, rectangle, minimum height=2em]
\tikzstyle{boxnode}=[boxnodewhite, fill=shadecolor]
\tikzstyle{boxnodehalf}=[boxnode, dashed, opacity=0.5]

\makeatletter
\newdimen\multi@col@width
\newdimen\multi@col@margin
\newcount\multi@col@count
\multi@col@width=0pt

\tikzset{
	multicol/.code={%
			\global\multi@col@count=#1\relax
			\global\let\orig@pgfmatrixendcode=\pgfmatrixendcode
			\global\let\orig@pgfmatrixemptycode=\pgfmatrixemptycode
			\def\pgfmatrixendcode##1{\orig@pgfmatrixendcode%
				##1%
				\pgfutil@tempdima=\pgf@picmaxx
				\global\multi@col@margin=\pgf@picminx
				\advance\pgfutil@tempdima by -\pgf@picminx
				\divide\pgfutil@tempdima by #1\relax
				\global\multi@col@width=\pgfutil@tempdima
				\pgf@picmaxx=.5\multi@col@width
				\pgf@picminx=-.5\multi@col@width
				\global\pgf@picmaxx=\pgf@picmaxx
				\global\pgf@picminx=\pgf@picminx
				\gdef\multi@adjust@position{%
					\setbox\pgf@matrix@cell=\hbox\bgroup
					\hfil\hskip-1.5\multi@col@margin
					\hfil\hskip-.5\multi@col@width
					\box\pgf@matrix@cell
					\egroup
				}%
				\gdef\multi@temp{\aftergroup\multi@adjust@position}%
				\aftergroup\multi@temp
			}
			\gdef\pgfmatrixemptycode{%
				\orig@pgfmatrixemptycode
				\global\advance\multi@col@count by -1\relax
				\global\pgf@picmaxx=.5\multi@col@width
				\global\pgf@picminx=-.5\multi@col@width
				\ifnum\multi@col@count=1\relax
					\global\let\pgfmatrixemptycode=\orig@pgfmatrixemptycode
				\fi
			}
		}
}
\makeatother

\tdplotsetmaincoords{78}{50}
\begin{tikzpicture}[
		tdplot_main_coords,
		layer/.style={draw opacity=1.0, fill opacity=0.1},
		layer0/.style={layer, draw=cLayer1, fill=cLayer1},
		layer1/.style={layer, draw=cLayer2, fill=cLayer2},
		layer2/.style={layer, draw=cLayer3, fill=cLayer3},
		layer3/.style={layer, draw=cLayer4, fill=cLayer4},
		layer4/.style={layer, draw=cLayer5, fill=cLayer5},
		layer5/.style={layer, draw=cLayer6, fill=cLayer6}
	]
	\def\layerdy{0.8}
	\def\layerdz{0.5}
    \def\layerdx{1}

	\def\layerwidth{8*\layerdx}
	\def\layerdepth{7*\layerdy}
	\def\layerheight{\layerdz}

    \foreach \x in {0,2,...,8}
    {
        \draw[layer0] (\x*\layerdx,0,0) -- (\x*\layerdx,\layerdepth,0);
    }
    \foreach \y in {0,1,...,7}
    {
        \draw[layer0] (0,\y*\layerdy,0) -- (\layerwidth,\y*\layerdy,0);
    }
	\foreach \z in {0,1,...,5} {
			\draw[layer\z] (0,0,\layerheight*\z) -- (\layerwidth,0,\layerheight*\z)
			-- (\layerwidth,\layerdepth,\layerheight*\z) -- (0,\layerdepth,\layerheight*\z) -- (0,0,\layerheight*\z) -- cycle;
			\foreach \layernum/\xstart/\ystart/\width/\depth in \highlightboxes
			{
				\ifnum\layernum=\z
					\pgfmathsetmacro{\zcoord}{\layerheight * \layernum}
					\pgfmathsetmacro{\boxheight}{1.0 * \layerdz}

					\coordinate (\ramiprefix-A) at (\xstart*\layerdx, \ystart*\layerdy, \zcoord );
					\coordinate (\ramiprefix-B) at (\xstart*\layerdx + \width*\layerdx, \ystart*\layerdy, \zcoord );
					\coordinate (\ramiprefix-C) at (\xstart*\layerdx + \width*\layerdx, \ystart*\layerdy + \depth*\layerdy, \zcoord );
					\coordinate (\ramiprefix-D) at (\xstart*\layerdx, \ystart*\layerdy + \depth*\layerdy, \zcoord );
					\coordinate (\ramiprefix-E) at (\xstart*\layerdx, \ystart*\layerdy, \zcoord + \boxheight );
					\coordinate (\ramiprefix-F) at (\xstart*\layerdx + \width*\layerdx, \ystart*\layerdy, \zcoord + \boxheight );
					\coordinate (\ramiprefix-G) at (\xstart*\layerdx + \width*\layerdx, \ystart*\layerdy + \depth*\layerdy, \zcoord + \boxheight );
					\coordinate (\ramiprefix-H) at (\xstart*\layerdx, \ystart*\layerdy + \depth*\layerdy, \zcoord + \boxheight );

					\draw[layer\layernum, fill opacity=0.6] (\ramiprefix-A) -- (\ramiprefix-B) -- (\ramiprefix-C) -- (\ramiprefix-D) -- cycle;
					\draw[layer\layernum, fill opacity=0.6] (\ramiprefix-E) -- (\ramiprefix-F) -- (\ramiprefix-G) -- (\ramiprefix-H) -- cycle;
					\draw[layer\layernum, fill opacity=0.6] (\ramiprefix-A) -- (\ramiprefix-B) -- (\ramiprefix-F) -- (\ramiprefix-E) -- cycle;
					\draw[layer\layernum, fill opacity=0.6] (\ramiprefix-B) -- (\ramiprefix-C) -- (\ramiprefix-G) -- (\ramiprefix-F) -- cycle;
					\draw[layer\layernum, fill opacity=0.6] (\ramiprefix-C) -- (\ramiprefix-D) -- (\ramiprefix-H) -- (\ramiprefix-G) -- cycle;
					\draw[layer\layernum, fill opacity=0.6] (\ramiprefix-A) -- (\ramiprefix-D) -- (\ramiprefix-H) -- (\ramiprefix-E) -- cycle;
				\fi
			}
		}

	\foreach \text/\step in {
			Asset/0, Integration/1, Communication/2,
			Information/3, Functional/4, Business/5
		}{
			\coordinate (\ramiprefix-trans-layers-\step) at (0,0,\step*\layerdz + 0.5 * \layerdz);

			\begin{scope}[
					shift=(\ramiprefix-trans-layers-\step),
					rotate around z=0,
					transform shape,
					canvas is xz plane at y=0
				]

				\draw (0,0,0) node[left] {\text};
			\end{scope}
		}

	\foreach \text/\step in {
			Product/0, Field Device/1, Control Device/2,
			Station/3, Work Centers/4, Enterprise/5,
			Connected World/6} {

			\coordinate (\ramiprefix-trans-hierarchy-\step) at (\layerwidth+0.1, \step*\layerdy + 0.5*\layerdy, -0.1);
			\begin{scope}[
					shift=(\ramiprefix-trans-hierarchy-\step),
					transform shape,
					canvas is xz plane at y=0
				]
				\draw (0,0,0) node[right] {\text};
			\end{scope}
		}

	\coordinate (\ramiprefix-trans-lifecycle) at (0, 0, -0.6);
	\begin{scope}[
			shift=(\ramiprefix-trans-lifecycle),
			transform shape,
			canvas is xz plane at y=0
		]
		\small
		\node[draw, fill=shadecolor, align=center,minimum width=2cm, minimum height=3em] at (1,0) {Development};
		\node[draw, fill=shadecolor, align=center,minimum width=2cm, minimum height=3em] at (3,0) {Mainten.\\ Usage};
		\node[draw, fill=shadecolor, align=center,minimum width=2cm, minimum height=3em] at (5,0) {Production};
		\node[draw, fill=shadecolor, align=center,minimum width=2cm, minimum height=3em] at (7,0) {Mainten.\\ Usage};
	\end{scope}

	\coordinate (\ramiprefix-trans-type) at (0, 0, -1.5);
	\begin{scope}[
			shift=(\ramiprefix-trans-type),
			transform shape,
			canvas is xz plane at y=0
		]
		\node[draw, fill=shadecolor, align=center, minimum width=4cm, minimum height=1.7em] at (2,0) {Type};
		\node[draw, fill=shadecolor, align=center, minimum width=4cm, minimum height=1.7em] at (6,0) {Instance};
	\end{scope}

	\coordinate (\ramiprefix-trans-lifecycle-arrow) at (0, 0, 5*\layerdz);
	\begin{scope}[
			shift=(\ramiprefix-trans-lifecycle-arrow),
			transform shape,
			canvas is xz plane at y=0,
		]

		\draw[->] (0,0) -- node[above, align=left]
		{IEC 62890\\ Life Cycle Value Stream}
		++(\layerwidth,0);
	\end{scope}

	\coordinate (\ramiprefix-trans-hierarchy-arrow) at (\layerwidth, 0, 5*\layerdz);
	\begin{scope}[
			shift=(\ramiprefix-trans-hierarchy-arrow),
			transform shape,
			canvas is yz plane at x=0
		]

		\draw[->] (0,0) -- node[above, align=left]
		{IEC 62264, IEC 61512\\ Hierarchy Levels}
		++(\layerdepth,0);
	\end{scope}

	\begin{scope}[
			transform shape,
			canvas is xz plane at y=0
		]

		\draw[->] (0,0) -- node[sloped, below] {Layers} ++(0, 5*\layerdz);
	\end{scope}

\end{tikzpicture}

\ifdefined\includingtikz
\else
\end{document}
\fi
		}
		\subcaption{Continuous Training}
		\label{fig:rami3}
		\undef\includingtikz
		\undef\ramiprefix
		\undef\highlightboxes
	\end{minipage}%
	\hfill
	\begin{minipage}{0.48\textwidth}
		\centering
		\def\includingtikz{1}
		\def\ramiprefix{A4}
		\def\highlightboxes{4/2/1/2/2,4/2/5/2/1}
		\scalebox{0.55}{
			\hspace{-2.5cm}
			\ifdefined\includingtikz
    \ifdefined\highlightboxes
    \else
        \def\highlightboxes{}
    \fi
\else
    \documentclass{standalone}
    \usepackage{graphicx}
    \usepackage{xcolor}
    \usepackage{tikz}
    \usepackage{tikz-3dplot}
    \usepackage{pgfplots}
    \pgfplotsset{compat=1.16}
    \def\highlightboxes{
			}
\begin{document}
\fi

\ifdefined\ramiprefix
\else
	\def\ramiprefix{X}
\fi

\definecolor{maincolor}{RGB}{35, 95, 157}
\definecolor{secondcolor}{RGB}{137, 157, 35}
\definecolor{alertcolor}{RGB}{192, 83, 84}
\definecolor{darkshadecolor}{RGB}{147,112,219}
\colorlet{shadecolor}{darkshadecolor!10}
\colorlet{shadegrey}{black!5}
\definecolor{cLayer1}{RGB}{72,209,204}
\definecolor{cLayer2}{RGB}{60,179,113}
\definecolor{cLayer3}{RGB}{255,140,0}
\definecolor{cLayer4}{RGB}{147,112,219}
\definecolor{cLayer5}{RGB}{205,92,92}
\definecolor{cLayer6}{RGB}{70,130,180}

\usetikzlibrary{calc, shapes.symbols, shapes.misc, shapes.arrows, decorations,
	decorations.pathmorphing, fit, matrix, matrix.skeleton, backgrounds}
\tikzset{>=latex}

\tikzstyle{boxnodewhite}=[draw, rectangle, minimum height=2em]
\tikzstyle{boxnode}=[boxnodewhite, fill=shadecolor]
\tikzstyle{boxnodehalf}=[boxnode, dashed, opacity=0.5]

\makeatletter
\newdimen\multi@col@width
\newdimen\multi@col@margin
\newcount\multi@col@count
\multi@col@width=0pt

\tikzset{
	multicol/.code={%
			\global\multi@col@count=#1\relax
			\global\let\orig@pgfmatrixendcode=\pgfmatrixendcode
			\global\let\orig@pgfmatrixemptycode=\pgfmatrixemptycode
			\def\pgfmatrixendcode##1{\orig@pgfmatrixendcode%
				##1%
				\pgfutil@tempdima=\pgf@picmaxx
				\global\multi@col@margin=\pgf@picminx
				\advance\pgfutil@tempdima by -\pgf@picminx
				\divide\pgfutil@tempdima by #1\relax
				\global\multi@col@width=\pgfutil@tempdima
				\pgf@picmaxx=.5\multi@col@width
				\pgf@picminx=-.5\multi@col@width
				\global\pgf@picmaxx=\pgf@picmaxx
				\global\pgf@picminx=\pgf@picminx
				\gdef\multi@adjust@position{%
					\setbox\pgf@matrix@cell=\hbox\bgroup
					\hfil\hskip-1.5\multi@col@margin
					\hfil\hskip-.5\multi@col@width
					\box\pgf@matrix@cell
					\egroup
				}%
				\gdef\multi@temp{\aftergroup\multi@adjust@position}%
				\aftergroup\multi@temp
			}
			\gdef\pgfmatrixemptycode{%
				\orig@pgfmatrixemptycode
				\global\advance\multi@col@count by -1\relax
				\global\pgf@picmaxx=.5\multi@col@width
				\global\pgf@picminx=-.5\multi@col@width
				\ifnum\multi@col@count=1\relax
					\global\let\pgfmatrixemptycode=\orig@pgfmatrixemptycode
				\fi
			}
		}
}
\makeatother

\tdplotsetmaincoords{78}{50}
\begin{tikzpicture}[
		tdplot_main_coords,
		layer/.style={draw opacity=1.0, fill opacity=0.1},
		layer0/.style={layer, draw=cLayer1, fill=cLayer1},
		layer1/.style={layer, draw=cLayer2, fill=cLayer2},
		layer2/.style={layer, draw=cLayer3, fill=cLayer3},
		layer3/.style={layer, draw=cLayer4, fill=cLayer4},
		layer4/.style={layer, draw=cLayer5, fill=cLayer5},
		layer5/.style={layer, draw=cLayer6, fill=cLayer6}
	]
	\def\layerdy{0.8}
	\def\layerdz{0.5}
    \def\layerdx{1}

	\def\layerwidth{8*\layerdx}
	\def\layerdepth{7*\layerdy}
	\def\layerheight{\layerdz}

    \foreach \x in {0,2,...,8}
    {
        \draw[layer0] (\x*\layerdx,0,0) -- (\x*\layerdx,\layerdepth,0);
    }
    \foreach \y in {0,1,...,7}
    {
        \draw[layer0] (0,\y*\layerdy,0) -- (\layerwidth,\y*\layerdy,0);
    }
	\foreach \z in {0,1,...,5} {
			\draw[layer\z] (0,0,\layerheight*\z) -- (\layerwidth,0,\layerheight*\z)
			-- (\layerwidth,\layerdepth,\layerheight*\z) -- (0,\layerdepth,\layerheight*\z) -- (0,0,\layerheight*\z) -- cycle;
			\foreach \layernum/\xstart/\ystart/\width/\depth in \highlightboxes
			{
				\ifnum\layernum=\z
					\pgfmathsetmacro{\zcoord}{\layerheight * \layernum}
					\pgfmathsetmacro{\boxheight}{1.0 * \layerdz}

					\coordinate (\ramiprefix-A) at (\xstart*\layerdx, \ystart*\layerdy, \zcoord );
					\coordinate (\ramiprefix-B) at (\xstart*\layerdx + \width*\layerdx, \ystart*\layerdy, \zcoord );
					\coordinate (\ramiprefix-C) at (\xstart*\layerdx + \width*\layerdx, \ystart*\layerdy + \depth*\layerdy, \zcoord );
					\coordinate (\ramiprefix-D) at (\xstart*\layerdx, \ystart*\layerdy + \depth*\layerdy, \zcoord );
					\coordinate (\ramiprefix-E) at (\xstart*\layerdx, \ystart*\layerdy, \zcoord + \boxheight );
					\coordinate (\ramiprefix-F) at (\xstart*\layerdx + \width*\layerdx, \ystart*\layerdy, \zcoord + \boxheight );
					\coordinate (\ramiprefix-G) at (\xstart*\layerdx + \width*\layerdx, \ystart*\layerdy + \depth*\layerdy, \zcoord + \boxheight );
					\coordinate (\ramiprefix-H) at (\xstart*\layerdx, \ystart*\layerdy + \depth*\layerdy, \zcoord + \boxheight );

					\draw[layer\layernum, fill opacity=0.6] (\ramiprefix-A) -- (\ramiprefix-B) -- (\ramiprefix-C) -- (\ramiprefix-D) -- cycle;
					\draw[layer\layernum, fill opacity=0.6] (\ramiprefix-E) -- (\ramiprefix-F) -- (\ramiprefix-G) -- (\ramiprefix-H) -- cycle;
					\draw[layer\layernum, fill opacity=0.6] (\ramiprefix-A) -- (\ramiprefix-B) -- (\ramiprefix-F) -- (\ramiprefix-E) -- cycle;
					\draw[layer\layernum, fill opacity=0.6] (\ramiprefix-B) -- (\ramiprefix-C) -- (\ramiprefix-G) -- (\ramiprefix-F) -- cycle;
					\draw[layer\layernum, fill opacity=0.6] (\ramiprefix-C) -- (\ramiprefix-D) -- (\ramiprefix-H) -- (\ramiprefix-G) -- cycle;
					\draw[layer\layernum, fill opacity=0.6] (\ramiprefix-A) -- (\ramiprefix-D) -- (\ramiprefix-H) -- (\ramiprefix-E) -- cycle;
				\fi
			}
		}

	\foreach \text/\step in {
			Asset/0, Integration/1, Communication/2,
			Information/3, Functional/4, Business/5
		}{
			\coordinate (\ramiprefix-trans-layers-\step) at (0,0,\step*\layerdz + 0.5 * \layerdz);

			\begin{scope}[
					shift=(\ramiprefix-trans-layers-\step),
					rotate around z=0,
					transform shape,
					canvas is xz plane at y=0
				]

				\draw (0,0,0) node[left] {\text};
			\end{scope}
		}

	\foreach \text/\step in {
			Product/0, Field Device/1, Control Device/2,
			Station/3, Work Centers/4, Enterprise/5,
			Connected World/6} {

			\coordinate (\ramiprefix-trans-hierarchy-\step) at (\layerwidth+0.1, \step*\layerdy + 0.5*\layerdy, -0.1);
			\begin{scope}[
					shift=(\ramiprefix-trans-hierarchy-\step),
					transform shape,
					canvas is xz plane at y=0
				]
				\draw (0,0,0) node[right] {\text};
			\end{scope}
		}

	\coordinate (\ramiprefix-trans-lifecycle) at (0, 0, -0.6);
	\begin{scope}[
			shift=(\ramiprefix-trans-lifecycle),
			transform shape,
			canvas is xz plane at y=0
		]
		\small
		\node[draw, fill=shadecolor, align=center,minimum width=2cm, minimum height=3em] at (1,0) {Development};
		\node[draw, fill=shadecolor, align=center,minimum width=2cm, minimum height=3em] at (3,0) {Mainten.\\ Usage};
		\node[draw, fill=shadecolor, align=center,minimum width=2cm, minimum height=3em] at (5,0) {Production};
		\node[draw, fill=shadecolor, align=center,minimum width=2cm, minimum height=3em] at (7,0) {Mainten.\\ Usage};
	\end{scope}

	\coordinate (\ramiprefix-trans-type) at (0, 0, -1.5);
	\begin{scope}[
			shift=(\ramiprefix-trans-type),
			transform shape,
			canvas is xz plane at y=0
		]
		\node[draw, fill=shadecolor, align=center, minimum width=4cm, minimum height=1.7em] at (2,0) {Type};
		\node[draw, fill=shadecolor, align=center, minimum width=4cm, minimum height=1.7em] at (6,0) {Instance};
	\end{scope}

	\coordinate (\ramiprefix-trans-lifecycle-arrow) at (0, 0, 5*\layerdz);
	\begin{scope}[
			shift=(\ramiprefix-trans-lifecycle-arrow),
			transform shape,
			canvas is xz plane at y=0,
		]

		\draw[->] (0,0) -- node[above, align=left]
		{IEC 62890\\ Life Cycle Value Stream}
		++(\layerwidth,0);
	\end{scope}

	\coordinate (\ramiprefix-trans-hierarchy-arrow) at (\layerwidth, 0, 5*\layerdz);
	\begin{scope}[
			shift=(\ramiprefix-trans-hierarchy-arrow),
			transform shape,
			canvas is yz plane at x=0
		]

		\draw[->] (0,0) -- node[above, align=left]
		{IEC 62264, IEC 61512\\ Hierarchy Levels}
		++(\layerdepth,0);
	\end{scope}

	\begin{scope}[
			transform shape,
			canvas is xz plane at y=0
		]

		\draw[->] (0,0) -- node[sloped, below] {Layers} ++(0, 5*\layerdz);
	\end{scope}

\end{tikzpicture}

\ifdefined\includingtikz
\else
\end{document}
\fi
		}
		\subcaption{Model Deployment}
		\label{fig:rami4}
		\undef\includingtikz
		\undef\ramiprefix
		\undef\highlightboxes
	\end{minipage}

	\vspace{2em}

	\begin{minipage}{0.48\textwidth}
		\centering
		\def\includingtikz{1}
		\def\ramiprefix{A5}
		\def\highlightboxes{0/4/1/2/5}
		\scalebox{0.55}{
			\hspace{-2.5cm}
			\ifdefined\includingtikz
    \ifdefined\highlightboxes
    \else
        \def\highlightboxes{}
    \fi
\else
    \documentclass{standalone}
    \usepackage{graphicx}
    \usepackage{xcolor}
    \usepackage{tikz}
    \usepackage{tikz-3dplot}
    \usepackage{pgfplots}
    \pgfplotsset{compat=1.16}
    \def\highlightboxes{
			}
\begin{document}
\fi

\ifdefined\ramiprefix
\else
	\def\ramiprefix{X}
\fi

\definecolor{maincolor}{RGB}{35, 95, 157}
\definecolor{secondcolor}{RGB}{137, 157, 35}
\definecolor{alertcolor}{RGB}{192, 83, 84}
\definecolor{darkshadecolor}{RGB}{147,112,219}
\colorlet{shadecolor}{darkshadecolor!10}
\colorlet{shadegrey}{black!5}
\definecolor{cLayer1}{RGB}{72,209,204}
\definecolor{cLayer2}{RGB}{60,179,113}
\definecolor{cLayer3}{RGB}{255,140,0}
\definecolor{cLayer4}{RGB}{147,112,219}
\definecolor{cLayer5}{RGB}{205,92,92}
\definecolor{cLayer6}{RGB}{70,130,180}

\usetikzlibrary{calc, shapes.symbols, shapes.misc, shapes.arrows, decorations,
	decorations.pathmorphing, fit, matrix, matrix.skeleton, backgrounds}
\tikzset{>=latex}

\tikzstyle{boxnodewhite}=[draw, rectangle, minimum height=2em]
\tikzstyle{boxnode}=[boxnodewhite, fill=shadecolor]
\tikzstyle{boxnodehalf}=[boxnode, dashed, opacity=0.5]

\makeatletter
\newdimen\multi@col@width
\newdimen\multi@col@margin
\newcount\multi@col@count
\multi@col@width=0pt

\tikzset{
	multicol/.code={%
			\global\multi@col@count=#1\relax
			\global\let\orig@pgfmatrixendcode=\pgfmatrixendcode
			\global\let\orig@pgfmatrixemptycode=\pgfmatrixemptycode
			\def\pgfmatrixendcode##1{\orig@pgfmatrixendcode%
				##1%
				\pgfutil@tempdima=\pgf@picmaxx
				\global\multi@col@margin=\pgf@picminx
				\advance\pgfutil@tempdima by -\pgf@picminx
				\divide\pgfutil@tempdima by #1\relax
				\global\multi@col@width=\pgfutil@tempdima
				\pgf@picmaxx=.5\multi@col@width
				\pgf@picminx=-.5\multi@col@width
				\global\pgf@picmaxx=\pgf@picmaxx
				\global\pgf@picminx=\pgf@picminx
				\gdef\multi@adjust@position{%
					\setbox\pgf@matrix@cell=\hbox\bgroup
					\hfil\hskip-1.5\multi@col@margin
					\hfil\hskip-.5\multi@col@width
					\box\pgf@matrix@cell
					\egroup
				}%
				\gdef\multi@temp{\aftergroup\multi@adjust@position}%
				\aftergroup\multi@temp
			}
			\gdef\pgfmatrixemptycode{%
				\orig@pgfmatrixemptycode
				\global\advance\multi@col@count by -1\relax
				\global\pgf@picmaxx=.5\multi@col@width
				\global\pgf@picminx=-.5\multi@col@width
				\ifnum\multi@col@count=1\relax
					\global\let\pgfmatrixemptycode=\orig@pgfmatrixemptycode
				\fi
			}
		}
}
\makeatother

\tdplotsetmaincoords{78}{50}
\begin{tikzpicture}[
		tdplot_main_coords,
		layer/.style={draw opacity=1.0, fill opacity=0.1},
		layer0/.style={layer, draw=cLayer1, fill=cLayer1},
		layer1/.style={layer, draw=cLayer2, fill=cLayer2},
		layer2/.style={layer, draw=cLayer3, fill=cLayer3},
		layer3/.style={layer, draw=cLayer4, fill=cLayer4},
		layer4/.style={layer, draw=cLayer5, fill=cLayer5},
		layer5/.style={layer, draw=cLayer6, fill=cLayer6}
	]
	\def\layerdy{0.8}
	\def\layerdz{0.5}
    \def\layerdx{1}

	\def\layerwidth{8*\layerdx}
	\def\layerdepth{7*\layerdy}
	\def\layerheight{\layerdz}

    \foreach \x in {0,2,...,8}
    {
        \draw[layer0] (\x*\layerdx,0,0) -- (\x*\layerdx,\layerdepth,0);
    }
    \foreach \y in {0,1,...,7}
    {
        \draw[layer0] (0,\y*\layerdy,0) -- (\layerwidth,\y*\layerdy,0);
    }
	\foreach \z in {0,1,...,5} {
			\draw[layer\z] (0,0,\layerheight*\z) -- (\layerwidth,0,\layerheight*\z)
			-- (\layerwidth,\layerdepth,\layerheight*\z) -- (0,\layerdepth,\layerheight*\z) -- (0,0,\layerheight*\z) -- cycle;
			\foreach \layernum/\xstart/\ystart/\width/\depth in \highlightboxes
			{
				\ifnum\layernum=\z
					\pgfmathsetmacro{\zcoord}{\layerheight * \layernum}
					\pgfmathsetmacro{\boxheight}{1.0 * \layerdz}

					\coordinate (\ramiprefix-A) at (\xstart*\layerdx, \ystart*\layerdy, \zcoord );
					\coordinate (\ramiprefix-B) at (\xstart*\layerdx + \width*\layerdx, \ystart*\layerdy, \zcoord );
					\coordinate (\ramiprefix-C) at (\xstart*\layerdx + \width*\layerdx, \ystart*\layerdy + \depth*\layerdy, \zcoord );
					\coordinate (\ramiprefix-D) at (\xstart*\layerdx, \ystart*\layerdy + \depth*\layerdy, \zcoord );
					\coordinate (\ramiprefix-E) at (\xstart*\layerdx, \ystart*\layerdy, \zcoord + \boxheight );
					\coordinate (\ramiprefix-F) at (\xstart*\layerdx + \width*\layerdx, \ystart*\layerdy, \zcoord + \boxheight );
					\coordinate (\ramiprefix-G) at (\xstart*\layerdx + \width*\layerdx, \ystart*\layerdy + \depth*\layerdy, \zcoord + \boxheight );
					\coordinate (\ramiprefix-H) at (\xstart*\layerdx, \ystart*\layerdy + \depth*\layerdy, \zcoord + \boxheight );

					\draw[layer\layernum, fill opacity=0.6] (\ramiprefix-A) -- (\ramiprefix-B) -- (\ramiprefix-C) -- (\ramiprefix-D) -- cycle;
					\draw[layer\layernum, fill opacity=0.6] (\ramiprefix-E) -- (\ramiprefix-F) -- (\ramiprefix-G) -- (\ramiprefix-H) -- cycle;
					\draw[layer\layernum, fill opacity=0.6] (\ramiprefix-A) -- (\ramiprefix-B) -- (\ramiprefix-F) -- (\ramiprefix-E) -- cycle;
					\draw[layer\layernum, fill opacity=0.6] (\ramiprefix-B) -- (\ramiprefix-C) -- (\ramiprefix-G) -- (\ramiprefix-F) -- cycle;
					\draw[layer\layernum, fill opacity=0.6] (\ramiprefix-C) -- (\ramiprefix-D) -- (\ramiprefix-H) -- (\ramiprefix-G) -- cycle;
					\draw[layer\layernum, fill opacity=0.6] (\ramiprefix-A) -- (\ramiprefix-D) -- (\ramiprefix-H) -- (\ramiprefix-E) -- cycle;
				\fi
			}
		}

	\foreach \text/\step in {
			Asset/0, Integration/1, Communication/2,
			Information/3, Functional/4, Business/5
		}{
			\coordinate (\ramiprefix-trans-layers-\step) at (0,0,\step*\layerdz + 0.5 * \layerdz);

			\begin{scope}[
					shift=(\ramiprefix-trans-layers-\step),
					rotate around z=0,
					transform shape,
					canvas is xz plane at y=0
				]

				\draw (0,0,0) node[left] {\text};
			\end{scope}
		}

	\foreach \text/\step in {
			Product/0, Field Device/1, Control Device/2,
			Station/3, Work Centers/4, Enterprise/5,
			Connected World/6} {

			\coordinate (\ramiprefix-trans-hierarchy-\step) at (\layerwidth+0.1, \step*\layerdy + 0.5*\layerdy, -0.1);
			\begin{scope}[
					shift=(\ramiprefix-trans-hierarchy-\step),
					transform shape,
					canvas is xz plane at y=0
				]
				\draw (0,0,0) node[right] {\text};
			\end{scope}
		}

	\coordinate (\ramiprefix-trans-lifecycle) at (0, 0, -0.6);
	\begin{scope}[
			shift=(\ramiprefix-trans-lifecycle),
			transform shape,
			canvas is xz plane at y=0
		]
		\small
		\node[draw, fill=shadecolor, align=center,minimum width=2cm, minimum height=3em] at (1,0) {Development};
		\node[draw, fill=shadecolor, align=center,minimum width=2cm, minimum height=3em] at (3,0) {Mainten.\\ Usage};
		\node[draw, fill=shadecolor, align=center,minimum width=2cm, minimum height=3em] at (5,0) {Production};
		\node[draw, fill=shadecolor, align=center,minimum width=2cm, minimum height=3em] at (7,0) {Mainten.\\ Usage};
	\end{scope}

	\coordinate (\ramiprefix-trans-type) at (0, 0, -1.5);
	\begin{scope}[
			shift=(\ramiprefix-trans-type),
			transform shape,
			canvas is xz plane at y=0
		]
		\node[draw, fill=shadecolor, align=center, minimum width=4cm, minimum height=1.7em] at (2,0) {Type};
		\node[draw, fill=shadecolor, align=center, minimum width=4cm, minimum height=1.7em] at (6,0) {Instance};
	\end{scope}

	\coordinate (\ramiprefix-trans-lifecycle-arrow) at (0, 0, 5*\layerdz);
	\begin{scope}[
			shift=(\ramiprefix-trans-lifecycle-arrow),
			transform shape,
			canvas is xz plane at y=0,
		]

		\draw[->] (0,0) -- node[above, align=left]
		{IEC 62890\\ Life Cycle Value Stream}
		++(\layerwidth,0);
	\end{scope}

	\coordinate (\ramiprefix-trans-hierarchy-arrow) at (\layerwidth, 0, 5*\layerdz);
	\begin{scope}[
			shift=(\ramiprefix-trans-hierarchy-arrow),
			transform shape,
			canvas is yz plane at x=0
		]

		\draw[->] (0,0) -- node[above, align=left]
		{IEC 62264, IEC 61512\\ Hierarchy Levels}
		++(\layerdepth,0);
	\end{scope}

	\begin{scope}[
			transform shape,
			canvas is xz plane at y=0
		]

		\draw[->] (0,0) -- node[sloped, below] {Layers} ++(0, 5*\layerdz);
	\end{scope}

\end{tikzpicture}

\ifdefined\includingtikz
\else
\end{document}
\fi
		}
		\subcaption{Prediction Serving}
		\label{fig:rami5}
		\undef\includingtikz
		\undef\ramiprefix
		\undef\highlightboxes
	\end{minipage}%
	\hfill
	\begin{minipage}{0.48\textwidth}
		\centering
		\def\includingtikz{1}
		\def\ramiprefix{A6}
		\def\highlightboxes{4/6/1/2/5,5/6/1/2/5}
		\scalebox{0.55}{
			\hspace{-2.5cm}
			\ifdefined\includingtikz
    \ifdefined\highlightboxes
    \else
        \def\highlightboxes{}
    \fi
\else
    \documentclass{standalone}
    \usepackage{graphicx}
    \usepackage{xcolor}
    \usepackage{tikz}
    \usepackage{tikz-3dplot}
    \usepackage{pgfplots}
    \pgfplotsset{compat=1.16}
    \def\highlightboxes{
			}
\begin{document}
\fi

\ifdefined\ramiprefix
\else
	\def\ramiprefix{X}
\fi

\definecolor{maincolor}{RGB}{35, 95, 157}
\definecolor{secondcolor}{RGB}{137, 157, 35}
\definecolor{alertcolor}{RGB}{192, 83, 84}
\definecolor{darkshadecolor}{RGB}{147,112,219}
\colorlet{shadecolor}{darkshadecolor!10}
\colorlet{shadegrey}{black!5}
\definecolor{cLayer1}{RGB}{72,209,204}
\definecolor{cLayer2}{RGB}{60,179,113}
\definecolor{cLayer3}{RGB}{255,140,0}
\definecolor{cLayer4}{RGB}{147,112,219}
\definecolor{cLayer5}{RGB}{205,92,92}
\definecolor{cLayer6}{RGB}{70,130,180}

\usetikzlibrary{calc, shapes.symbols, shapes.misc, shapes.arrows, decorations,
	decorations.pathmorphing, fit, matrix, matrix.skeleton, backgrounds}
\tikzset{>=latex}

\tikzstyle{boxnodewhite}=[draw, rectangle, minimum height=2em]
\tikzstyle{boxnode}=[boxnodewhite, fill=shadecolor]
\tikzstyle{boxnodehalf}=[boxnode, dashed, opacity=0.5]

\makeatletter
\newdimen\multi@col@width
\newdimen\multi@col@margin
\newcount\multi@col@count
\multi@col@width=0pt

\tikzset{
	multicol/.code={%
			\global\multi@col@count=#1\relax
			\global\let\orig@pgfmatrixendcode=\pgfmatrixendcode
			\global\let\orig@pgfmatrixemptycode=\pgfmatrixemptycode
			\def\pgfmatrixendcode##1{\orig@pgfmatrixendcode%
				##1%
				\pgfutil@tempdima=\pgf@picmaxx
				\global\multi@col@margin=\pgf@picminx
				\advance\pgfutil@tempdima by -\pgf@picminx
				\divide\pgfutil@tempdima by #1\relax
				\global\multi@col@width=\pgfutil@tempdima
				\pgf@picmaxx=.5\multi@col@width
				\pgf@picminx=-.5\multi@col@width
				\global\pgf@picmaxx=\pgf@picmaxx
				\global\pgf@picminx=\pgf@picminx
				\gdef\multi@adjust@position{%
					\setbox\pgf@matrix@cell=\hbox\bgroup
					\hfil\hskip-1.5\multi@col@margin
					\hfil\hskip-.5\multi@col@width
					\box\pgf@matrix@cell
					\egroup
				}%
				\gdef\multi@temp{\aftergroup\multi@adjust@position}%
				\aftergroup\multi@temp
			}
			\gdef\pgfmatrixemptycode{%
				\orig@pgfmatrixemptycode
				\global\advance\multi@col@count by -1\relax
				\global\pgf@picmaxx=.5\multi@col@width
				\global\pgf@picminx=-.5\multi@col@width
				\ifnum\multi@col@count=1\relax
					\global\let\pgfmatrixemptycode=\orig@pgfmatrixemptycode
				\fi
			}
		}
}
\makeatother

\tdplotsetmaincoords{78}{50}
\begin{tikzpicture}[
		tdplot_main_coords,
		layer/.style={draw opacity=1.0, fill opacity=0.1},
		layer0/.style={layer, draw=cLayer1, fill=cLayer1},
		layer1/.style={layer, draw=cLayer2, fill=cLayer2},
		layer2/.style={layer, draw=cLayer3, fill=cLayer3},
		layer3/.style={layer, draw=cLayer4, fill=cLayer4},
		layer4/.style={layer, draw=cLayer5, fill=cLayer5},
		layer5/.style={layer, draw=cLayer6, fill=cLayer6}
	]
	\def\layerdy{0.8}
	\def\layerdz{0.5}
    \def\layerdx{1}

	\def\layerwidth{8*\layerdx}
	\def\layerdepth{7*\layerdy}
	\def\layerheight{\layerdz}

    \foreach \x in {0,2,...,8}
    {
        \draw[layer0] (\x*\layerdx,0,0) -- (\x*\layerdx,\layerdepth,0);
    }
    \foreach \y in {0,1,...,7}
    {
        \draw[layer0] (0,\y*\layerdy,0) -- (\layerwidth,\y*\layerdy,0);
    }
	\foreach \z in {0,1,...,5} {
			\draw[layer\z] (0,0,\layerheight*\z) -- (\layerwidth,0,\layerheight*\z)
			-- (\layerwidth,\layerdepth,\layerheight*\z) -- (0,\layerdepth,\layerheight*\z) -- (0,0,\layerheight*\z) -- cycle;
			\foreach \layernum/\xstart/\ystart/\width/\depth in \highlightboxes
			{
				\ifnum\layernum=\z
					\pgfmathsetmacro{\zcoord}{\layerheight * \layernum}
					\pgfmathsetmacro{\boxheight}{1.0 * \layerdz}

					\coordinate (\ramiprefix-A) at (\xstart*\layerdx, \ystart*\layerdy, \zcoord );
					\coordinate (\ramiprefix-B) at (\xstart*\layerdx + \width*\layerdx, \ystart*\layerdy, \zcoord );
					\coordinate (\ramiprefix-C) at (\xstart*\layerdx + \width*\layerdx, \ystart*\layerdy + \depth*\layerdy, \zcoord );
					\coordinate (\ramiprefix-D) at (\xstart*\layerdx, \ystart*\layerdy + \depth*\layerdy, \zcoord );
					\coordinate (\ramiprefix-E) at (\xstart*\layerdx, \ystart*\layerdy, \zcoord + \boxheight );
					\coordinate (\ramiprefix-F) at (\xstart*\layerdx + \width*\layerdx, \ystart*\layerdy, \zcoord + \boxheight );
					\coordinate (\ramiprefix-G) at (\xstart*\layerdx + \width*\layerdx, \ystart*\layerdy + \depth*\layerdy, \zcoord + \boxheight );
					\coordinate (\ramiprefix-H) at (\xstart*\layerdx, \ystart*\layerdy + \depth*\layerdy, \zcoord + \boxheight );

					\draw[layer\layernum, fill opacity=0.6] (\ramiprefix-A) -- (\ramiprefix-B) -- (\ramiprefix-C) -- (\ramiprefix-D) -- cycle;
					\draw[layer\layernum, fill opacity=0.6] (\ramiprefix-E) -- (\ramiprefix-F) -- (\ramiprefix-G) -- (\ramiprefix-H) -- cycle;
					\draw[layer\layernum, fill opacity=0.6] (\ramiprefix-A) -- (\ramiprefix-B) -- (\ramiprefix-F) -- (\ramiprefix-E) -- cycle;
					\draw[layer\layernum, fill opacity=0.6] (\ramiprefix-B) -- (\ramiprefix-C) -- (\ramiprefix-G) -- (\ramiprefix-F) -- cycle;
					\draw[layer\layernum, fill opacity=0.6] (\ramiprefix-C) -- (\ramiprefix-D) -- (\ramiprefix-H) -- (\ramiprefix-G) -- cycle;
					\draw[layer\layernum, fill opacity=0.6] (\ramiprefix-A) -- (\ramiprefix-D) -- (\ramiprefix-H) -- (\ramiprefix-E) -- cycle;
				\fi
			}
		}

	\foreach \text/\step in {
			Asset/0, Integration/1, Communication/2,
			Information/3, Functional/4, Business/5
		}{
			\coordinate (\ramiprefix-trans-layers-\step) at (0,0,\step*\layerdz + 0.5 * \layerdz);

			\begin{scope}[
					shift=(\ramiprefix-trans-layers-\step),
					rotate around z=0,
					transform shape,
					canvas is xz plane at y=0
				]

				\draw (0,0,0) node[left] {\text};
			\end{scope}
		}

	\foreach \text/\step in {
			Product/0, Field Device/1, Control Device/2,
			Station/3, Work Centers/4, Enterprise/5,
			Connected World/6} {

			\coordinate (\ramiprefix-trans-hierarchy-\step) at (\layerwidth+0.1, \step*\layerdy + 0.5*\layerdy, -0.1);
			\begin{scope}[
					shift=(\ramiprefix-trans-hierarchy-\step),
					transform shape,
					canvas is xz plane at y=0
				]
				\draw (0,0,0) node[right] {\text};
			\end{scope}
		}

	\coordinate (\ramiprefix-trans-lifecycle) at (0, 0, -0.6);
	\begin{scope}[
			shift=(\ramiprefix-trans-lifecycle),
			transform shape,
			canvas is xz plane at y=0
		]
		\small
		\node[draw, fill=shadecolor, align=center,minimum width=2cm, minimum height=3em] at (1,0) {Development};
		\node[draw, fill=shadecolor, align=center,minimum width=2cm, minimum height=3em] at (3,0) {Mainten.\\ Usage};
		\node[draw, fill=shadecolor, align=center,minimum width=2cm, minimum height=3em] at (5,0) {Production};
		\node[draw, fill=shadecolor, align=center,minimum width=2cm, minimum height=3em] at (7,0) {Mainten.\\ Usage};
	\end{scope}

	\coordinate (\ramiprefix-trans-type) at (0, 0, -1.5);
	\begin{scope}[
			shift=(\ramiprefix-trans-type),
			transform shape,
			canvas is xz plane at y=0
		]
		\node[draw, fill=shadecolor, align=center, minimum width=4cm, minimum height=1.7em] at (2,0) {Type};
		\node[draw, fill=shadecolor, align=center, minimum width=4cm, minimum height=1.7em] at (6,0) {Instance};
	\end{scope}

	\coordinate (\ramiprefix-trans-lifecycle-arrow) at (0, 0, 5*\layerdz);
	\begin{scope}[
			shift=(\ramiprefix-trans-lifecycle-arrow),
			transform shape,
			canvas is xz plane at y=0,
		]

		\draw[->] (0,0) -- node[above, align=left]
		{IEC 62890\\ Life Cycle Value Stream}
		++(\layerwidth,0);
	\end{scope}

	\coordinate (\ramiprefix-trans-hierarchy-arrow) at (\layerwidth, 0, 5*\layerdz);
	\begin{scope}[
			shift=(\ramiprefix-trans-hierarchy-arrow),
			transform shape,
			canvas is yz plane at x=0
		]

		\draw[->] (0,0) -- node[above, align=left]
		{IEC 62264, IEC 61512\\ Hierarchy Levels}
		++(\layerdepth,0);
	\end{scope}

	\begin{scope}[
			transform shape,
			canvas is xz plane at y=0
		]

		\draw[->] (0,0) -- node[sloped, below] {Layers} ++(0, 5*\layerdz);
	\end{scope}

\end{tikzpicture}

\ifdefined\includingtikz
\else
\end{document}
\fi
		}
		\subcaption{Continuous Monitoring}
		\label{fig:rami6}
		\undef\includingtikz
		\undef\ramiprefix
		\undef\highlightboxes
	\end{minipage}

	\caption{Embedding of the \gls{MLOps} lifecycle into the \gls{R4} model.}
	\label{fig:rami-collection}
\end{figure}

\section{MLOps on an ICS Testbed} This section describes the setup of a
\gls{MLOps} pipeline on a testbed simulating an \gls{ICS} environment and evaluates
the mapping of \gls{MLOps} to the \gls{R4} model using the \gls{R4} Toolbox~\cite{binder2021towards} and
system models based on \gls{MBSE}.

\subsection{Testbed Description}
For our evaluation, we utilize an industrial testbed designed to simulate a
whole production process including three injection molding machines, four
robotic arms, a conveyor belt, and the necessary encompassing multi-layered
architecture. It reflects a real industrial environment, with distinct \gls{OT}
and \gls{IT} domains separated by appropriate security boundaries.

The core automation layer consists of multiple PLCs (B\&R and Sigmatek) that
control simulated physical processes, while Automation PCs running KUKA
software manage robotic operations. A \gls{SCADA} system implemented with
COPA-DATA Zenon \sthu{Is it ZENON or Zenon? We should also name COPA-DATA, so
	it is clear.} provides supervisory control and visualization. Network
infrastructure includes segmented switches and firewalls enforcing zone
separation according to ISO 62443 and more industrial security best practices,
including granular access control and a separated zone for insecure legacy
devices.
\sthu{Maybe ask Thomas Rosenstatter, but we do not only follow best practices,
	but ISO 62443.}

The data flow begins with process data generated by PLCs and exposed via \gls{OPCUA}
servers. A central \gls{IE} collects this data and
forwards it to a Kafka cluster for streaming and transformation. This
architecture enables data to be processed, stored, and made available to
analytics systems while maintaining \gls{OT} operational integrity. Data
storage follows a tiered approach with short-term retention in the \gls{IE} and
long-term archival in NAS infrastructure.

For \gls{ml}, the testbed connects an on-premises dedicated GPU cluster using SLURM
for computational resource management with an on-premises MLFlow instance that
tracks training runs, model versions, and enables deployment to edge devices as
well as the GPU cluster for prediction serving. This setup allows
\gls{ml} workloads to be executed from within the \gls{OT} environment in
real-time. By monitoring the performance of the deployed models through the
\gls{IE} and using the acquired insights to improve existing and develop new
models, this design creates a complete \gls{MLOps} cycle while respecting the
constraints and requirements of \glspl{ICS}.

\subsection{Evaluation of the mapping}
This section evaluates the \gls{MLOps}
to \gls{R4} mapping (\cref{fig:rami-collection}) using the \gls{R4} Toolbox and \gls{MBSE}-based system
models following \gls{R4} principals, with emphasis on the dynamic aspects
demonstrated in our testbed use case. While the application of the \gls{R4} Toolbox
and the accompanied methodology of creating system models is precisely described
in~\cite{binder2021towards}, this work makes use of the previous introduced concepts.
However, creating such models help to evaluate the proposed interconnection between
\gls{R4} and \gls{MLOps} by implementing an actual example as Proof-of-Concept (PoC)
within already established engineering methodologies.

Data ingestion within the \gls{MLOps} \textbf{Data and Model Management} phase is mapped
to the \gls{R4} Asset and Integration Layers, where \gls{SysML} models provide
formal representation of physical components. This addresses the \enquote{Data
	and Integration Requirements} challenge by standardizing heterogeneous data
sources. The \gls{R4} Toolbox~\cite{binder2021towards} links \gls{SysML} models with the \gls{AAS},
enabling structured data input from real-world components. The testbed
demonstrates how PLCs gather and manage production system data, overcoming
\enquote{Heterogeneity and Legacy Equipment} challenges.

The \textbf{Training Operationalization} Phase consisting of Data preprocessing
and training is situated in \gls{R4}'s Information Layer,
utilizing \glspl{DSL} for data clustering and storage, with data flow diagrams
defining transformation processes. Placing computationally intensive operations
at higher hierarchy levels addresses \enquote{Infrastructure and Resource
	Constraints}. The evaluation shows how the GPU cluster with SLURM utilizes
specific datasets and interacts with the architecture via \gls{OPCUA}, ensuring
properly formatted data for model training while maintaining separation from
safety-critical systems.

\gls{ml} \textbf{Model Development} occurs in specialized tools, with integration
represented in the Information and Communication Layers to facilitate data flow
and interoperability. The testbed's GPU cluster efficiently executes these
computationally intensive tasks.

\textbf{Continuous Training} relies on data collection from the central \gls{IE},
addressing the \enquote{Validation and Testing Constraints} by enabling
systematic improvement without operational disruption. The Kafka cluster feeds
data back into the processing pipeline, allowing iterative model optimization
and maintaining high adaptability within the production environment.

\textbf{Model Deployment} is modeled in \gls{R4}'s Communication Layer, linking
trained models to technical components like edge devices or inference servers.
This addresses \enquote{Real-time and Deterministic Requirements} by optimizing
inference placement. The testbed demonstrates how models are deployed to
specific hardware and interface with system infrastructure, maintaining
\enquote{High Reliability and Availability} requirements.

\textbf{Prediction Serving} in production is facilitated through the central
\gls{IE}, addressing \enquote{Architectural Considerations} by respecting
existing hierarchies while introducing \gls{ml} capabilities. \gls{R4}'s
holistic representation enables monitoring of applied models, demonstrating
\gls{MLOps}' practical impact in industrial automation.

\textbf{Continuous Monitoring} aligns with \gls{R4}'s Instance/Usage
perspective, addressing \enquote{Safety-Critical and Regulated Environments}
challenges by verifying model performance against safety requirements. The
testbed's components interconnect via \gls{OPCUA}, enabling feedback loops that
support model refinement while maintaining regulatory compliance. While this
phase is not directly modeled within \gls{R4}, tool-chain traceability ensures
that runtime monitoring has strong dependencies on prior \gls{R4} models.

By leveraging the \gls{R4} Toolbox~\cite{binder2021towards} and \gls{MBSE}-based system models, this
evaluation provides a structured approach to integrating \gls{MLOps} within
\gls{R4}, ensuring a comprehensive framework for data-driven industrial
automation that systematically addresses the challenges outlined in
\cref{sec:challenges}.
%
\section{Discussion and Conclusion} \label{sec:discussion-conclusion}
Our research demonstrates that embedding the \gls{MLOps} lifecycle into
\gls{OT} reference models provides a structured approach for integrating
\gls{ml} capabilities into industrial environments. While direct
transplantation of standard \gls{MLOps} practices to \gls{OT} environments is
challenging, our mapping to \gls{R4} offers a viable integration pathway.

For data scientists, \gls{OT} systems present significant structural
complexity. Reference models like \gls{R4} reduce this complexity by making
common patterns and features explicit. By embedding \gls{MLOps} into \gls{R4},
our work systematizes \gls{MLOps} practices for \gls{OT} systems while
maintaining adherence to industrial requirements.

This mapping addresses key challenges by: providing structured data management
that respects system heterogeneity; establishing clear computational resource
boundaries; creating well-defined model deployment pathways that maintain
reliability; and enabling continuous monitoring within safety and regulatory
constraints.

Another critical finding of our work is the value of controlled environments
for testing and validating \gls{ml} systems in industrial settings. The
presented testbed proved invaluable for evaluating our \gls{ml} approaches,
security strategies and architectural considerations like this systematic
embedding of \gls{MLOps} into \gls{R4}. It provides a realistic yet safe
environment that simulates complex industrial processes while allowing for
experimentation without risking operational disruptions.

Future work should expand evaluation to additional industrial use cases and
explore sector-specific adaptations. Further research into model
validation within \gls{OT} constraints would enhance practical applicability.

In conclusion, our systematic embedding of the \gls{MLOps} lifecycle into
\gls{R4} provides a blueprint for integrating \gls{ml} capabilities into
industrial environments while maintaining operational stability, reducing
complexity and systematizing implementation to make use of \gls{ml} within
\gls{OT} constraints.

\ifNotBlindReview
	\section*{Acknowledgment}
	The financial support by the Christian Doppler Research Association, the
	Austrian Federal Ministry for Digital and Economic Affairs and the Federal
	State of Salzburg is gratefully acknowledged.
\fi
\printbibliography

\end{document}